\documentclass{article}
\PassOptionsToPackage{numbers, compress}{natbib}

\usepackage[final]{neurips_2022}

\usepackage[utf8]{inputenc} 
\usepackage[T1]{fontenc}    
\usepackage[pagebackref,breaklinks,colorlinks]{hyperref}
\usepackage{url}            
\usepackage{booktabs}       
\usepackage{amsfonts}       
\usepackage{nicefrac}       
\usepackage{microtype}      
\usepackage{xcolor}         
\usepackage{longtable}
\usepackage{nicefrac}       
\usepackage{microtype}      
\usepackage{multirow}       
\usepackage{wrapfig}
\usepackage{pifont}
\usepackage[multiple]{footmisc}
\usepackage{makecell}
\usepackage{caption}
\usepackage{adjustbox}
\usepackage{diagbox} 
\usepackage{changepage}
\usepackage{makecell}
\usepackage{subfigure}
\usepackage{caption}
\usepackage{arydshln}
\usepackage{letltxmacro}
\usepackage{xspace}
\usepackage{bm}
\usepackage[ruled,linesnumbered,lined]{algorithm2e}

\usepackage{enumerate}
\title{CASA: Category-agnostic Skeletal Animal Reconstruction}

\author{%
Yuefan Wu$^1$\thanks{Indicates equal contribution}\hspace{4mm}
Zeyuan Chen$^{1\ast}$\hspace{4mm}
Shaowei Liu$^2$\hspace{4mm} 
Zhongzheng Ren$^2$\hspace{4mm} 
Shenlong Wang$^2$\hspace{4mm} \\
$^1$ University of Science and Technology of China\qquad $^2$University of Illinois Urbana-Champaign\\
{\normalsize \url{https://Iven-Wu.github.io/CASA}}
}

\begin{document}
\newcommand{\Perp}{\perp\!\!\! \perp}


\def\hpY{\mathbf{\bar{\beta}}}

\newcommand{\todo}[1]{{\color{red}{TODO: #1}}}
\newcommand\shenlong[1]{\textcolor{magenta}{#1}}
\newcommand\shenlongcomment[1]{\textcolor{magenta}{Shenlong: #1}}
\newcommand{\jr}[1]{{\color{orange}{#1}}}
\newcommand\jrc[1]{\textcolor{orange}{JR: #1}}

\newcommand\shaowei[1]{\textcolor{green}{Shaowei: #1}}

\newcolumntype{H}{>{\setbox0=\hbox\bgroup}c<{\egroup}@{}}

\newcommand{\lidar}{LiDAR}
\newcommand{\dataset}{PlanetZoo}

\definecolor{purple}{RGB}{160, 32, 240}
\definecolor{washblue}{RGB}{186, 224, 228}
\definecolor{sky}{RGB}{128, 128, 128}
\definecolor{seagreen}{RGB}{60, 179, 113}
\definecolor{building}{RGB}{128, 0, 0}
\definecolor{road}{RGB}{128, 64, 128}
\definecolor{sidewalk}{RGB}{0, 0, 192}
\definecolor{fence}{RGB}{64, 64, 128}
\definecolor{vegetation}{RGB}{128, 128, 0}
\definecolor{car}{RGB}{64, 0, 128}
\definecolor{sign}{RGB}{192, 128, 128}
\definecolor{pedestrian}{RGB}{64, 64, 0}
\definecolor{cyclist}{RGB}{0, 128, 192}

\def\be {\begin{equation}}
\def\ee {\end{equation}}
\def\beas {\begin{eqnarray*}}
\def\eeas {\end{eqnarray*}}
\def\bea {\begin{eqnarray}}
\def\eea {\end{eqnarray}}
\def\bes {\begin{equation*}}
\def\ees {\end{equation*}}

\newcommand{\fref}[1]{Fig.~\ref{#1}}
\newcommand{\tref}[1]{Table~\ref{#1}}
\newcommand{\Sref}[1]{Section~\ref{#1}}
\newcommand{\aref}[1]{Algorithm~\ref{#1}}

\newcommand{\figref}[1]{Fig\onedot~\ref{#1}}
\newcommand{\equref}[1]{Eq\onedot~\ref{#1}}
\newcommand{\secref}[1]{Sec\onedot~\ref{#1}}
\newcommand{\tabref}[1]{Tab\onedot~\ref{#1}}
\newcommand{\thmref}[1]{Theorem~\ref{#1}}
\newcommand{\prgref}[1]{Program~\ref{#1}}
\newcommand{\algref}[1]{Alg\onedot~\ref{#1}}
\newcommand{\clmref}[1]{Claim~\ref{#1}}
\newcommand{\lemref}[1]{Lemma~\ref{#1}}
\newcommand{\ptyref}[1]{Property~\ref{#1}}
\newcommand{\propref}[1]{Proposition~\ref{#1}}

\newcommand{\bbR}{{\mathbb{R}}}
\newcommand{\bK}{\mathbf{K}}
\newcommand{\bX}{\mathbf{X}}
\newcommand{\bY}{\mathbf{Y}}
\newcommand{\bk}{\mathbf{k}}
\newcommand{\bx}{\mathbf{x}}
\newcommand{\by}{\mathbf{y}}
\newcommand{\bhy}{\hat{\mathbf{y}}}
\newcommand{\bty}{\tilde{\mathbf{y}}}
\newcommand{\bG}{\mathbf{G}}
\newcommand{\bI}{\mathbf{I}}
\newcommand{\bg}{\mathbf{g}}
\newcommand{\bS}{\mathbf{S}}
\newcommand{\bs}{\mathbf{s}}
\newcommand{\bM}{\mathbf{M}}
\newcommand{\bw}{\mathbf{w}}
\newcommand{\eye}{\mathbf{I}}
\newcommand{\bU}{\mathbf{U}}
\newcommand{\bV}{\mathbf{V}}
\newcommand{\bW}{\mathbf{W}}
\newcommand{\bn}{\mathbf{n}}
\newcommand{\bv}{\mathbf{v}}
\newcommand{\bq}{\mathbf{q}}
\newcommand{\bR}{\mathbf{R}}
\newcommand{\bi}{\mathbf{i}}
\newcommand{\bj}{\mathbf{j}}
\newcommand{\bp}{\mathbf{p}}
\newcommand{\bt}{\mathbf{t}}
\newcommand{\bJ}{\mathbf{J}}
\newcommand{\bu}{\mathbf{u}}
\newcommand{\bB}{\mathbf{B}}
\newcommand{\bD}{\mathbf{D}}
\newcommand{\bz}{\mathbf{z}}
\newcommand{\bP}{\mathbf{P}}
\newcommand{\bC}{\mathbf{C}}
\newcommand{\bA}{\mathbf{A}}
\newcommand{\bZ}{\mathbf{Z}}
\newcommand{\bff}{\mathbf{f}}
\newcommand{\bF}{\mathbf{F}}
\newcommand{\bo}{\mathbf{o}}
\newcommand{\bc}{\mathbf{c}}
\newcommand{\bT}{\mathbf{T}}
\newcommand{\bQ}{\mathbf{Q}}
\newcommand{\bL}{\mathbf{L}}
\newcommand{\bl}{\mathbf{l}}
\newcommand{\ba}{\mathbf{a}}
\newcommand{\bE}{\mathbf{E}}
\newcommand{\bH}{\mathbf{H}}
\newcommand{\bd}{\mathbf{d}}
\newcommand{\br}{\mathbf{r}}
\newcommand{\bb}{\mathbf{b}}
\newcommand{\bh}{\mathbf{h}}

\newcommand{\btheta}{\bm{\theta}}
\newcommand{\bhh}{\hat{\mathbf{h}}}
\newcommand{\ci}{{\cal I}}
\newcommand{\ct}{{\cal T}}
\newcommand{\co}{{\cal O}}
\newcommand{\ck}{{\cal K}}
\newcommand{\cu}{{\cal U}}
\newcommand{\cS}{{\cal S}}
\newcommand{\cQ}{{\cal Q}}
\newcommand{\cT}{{\cal S}}
\newcommand{\cC}{{\cal C}}
\newcommand{\cE}{{\cal E}}
\newcommand{\cF}{{\cal F}}
\newcommand{\cL}{{\cal L}}
\newcommand{\X}{{\cal{X}}}
\newcommand{\Y}{{\cal Y}}
\newcommand{\cH}{{\cal H}}
\newcommand{\cP}{{\cal P}}
\newcommand{\cN}{{\cal N}}
\newcommand{\cU}{{\cal U}}
\newcommand{\cV}{{\cal V}}
\newcommand{\cX}{{\cal X}}
\newcommand{\cY}{{\cal Y}}
\newcommand{\graph}{{\cal H}}
\newcommand{\bayes}{{\cal B}}
\newcommand{\cx}{{\cal X}}
\newcommand{\cg}{{\cal G}}
\newcommand{\cm}{{\cal M}}
\newcommand{\cM}{{\cal M}}
\newcommand{\cG}{{\cal G}}
\newcommand{\cR}{\cal{R}}
\newcommand{\R}{\cal{R}}
\newcommand{\eig}{\mathrm{eig}}

\newcommand{\D}{{\cal D}}
\newcommand{\bfp}{{\bf p}}
\newcommand{\bfd}{{\bf d}}

\newcommand{\cv}{{\cal V}}
\newcommand{\ce}{{\cal E}}
\newcommand{\cy}{{\cal Y}}
\newcommand{\cz}{{\cal Z}}
\newcommand{\cb}{{\cal B}}
\newcommand{\cq}{{\cal Q}}
\newcommand{\cd}{{\cal D}}
\newcommand{\bcf}{{\cal F}}
\newcommand{\cI}{\mathcal{I}}

\newcommand{\ut}{^{(t)}}
\newcommand{\up}{^{(t-1)}}

\newcommand{\bpi}{\boldsymbol{\pi}}
\newcommand{\bphi}{\boldsymbol{\phi}}
\newcommand{\bPhi}{\boldsymbol{\Phi}}
\newcommand{\bmu}{\boldsymbol{\mu}}
\newcommand{\bSigma}{\boldsymbol{\Sigma}}
\newcommand{\bGamma}{\boldsymbol{\Gamma}}
\newcommand{\bbeta}{\boldsymbol{\beta}}
\newcommand{\bomega}{\boldsymbol{\omega}}
\newcommand{\blambda}{\boldsymbol{\lambda}}
\newcommand{\bkappa}{\boldsymbol{\kappa}}
\newcommand{\btau}{\boldsymbol{\tau}}
\newcommand{\balpha}{\boldsymbol{\alpha}}
\def\bgamma{\boldsymbol\gamma}

\newcommand{\prox}{{\mathrm{prox}}}

\newcommand{\pardev}[2]{\frac{\partial #1}{\partial #2}}
\newcommand{\dev}[2]{\frac{d #1}{d #2}}
\newcommand{\dw}{\delta\bw}
\newcommand{\lab}{\mathcal{L}}
\newcommand{\unlab}{\mathcal{U}}
\newcommand{\ind}{1{\hskip -2.5 pt}\hbox{I}}
\newcommand{\ff}[2]{   \cf_{\prec (#1 \rightarrow #2)}}
\newcommand{\vv}[2]{   \cv_{\prec (#1 \rightarrow #2)}}
\newcommand{\dd}[2]{   \delta_{#1 \rightarrow #2}}
\newcommand{\ld}[2]{   \lambda_{#1 \rightarrow #2}}
\newcommand{\en}[2]{  \bD(#1|| #2)}
\newcommand{\ex}[3]{  \bE_{#1 \sim #2}\left[ #3\right]} 
\newcommand{\exd}[2]{  \bE_{#1 }\left[ #2\right]}

\def\eop {{\noindent\framebox[0.5em]{\rule[0.25ex]{0em}{0.75ex}}}}

\newcommand{\tr}[1]{\ensuremath{\mathrm{tr}\left(#1\right)}}
\def\Xdim{{d}}
\def\Ydim{{D}}
\def\Zdim{{S}}

\def\tran {\top}

\makeatletter
\def\@onedot{\ifx\@let@token.\else.\null\fi\xspace}
\DeclareRobustCommand\onedot{\futurelet\@let@token\@onedot}

\def\eg{\emph{e.g}\onedot} \def\Eg{\emph{E.g}\onedot}
\def\ie{\emph{i.e}\onedot} \def\Ie{\emph{I.e}\onedot}
\def\cf{\emph{c.f}\onedot} \def\Cf{\emph{C.f}\onedot}
\def\etc{\emph{etc}\onedot} \def\vs{\emph{vs}\onedot}
\def\wrt{w.r.t\onedot} \def\dof{d.o.f\onedot}
\def\etal{\emph{et al}\onedot}

\newcommand{\se}[1]{\mathfrak{se}(#1)}
\newcommand{\SE}[1]{\mathbb{SE}(#1)}
\newcommand{\so}[1]{\mathfrak{so}(#1)}
\newcommand{\SO}[1]{\mathbb{SO}(#1)}

\newcommand{\poselow}{\xi}
\newcommand{\pose}{\bm{\poselow}}
\newcommand{\linpose}{\pose^\ell}
\newcommand{\cbpose}{\pose^c}
\newcommand{\rateparam}{v_i}
\newcommand{\bapose}{\bm{\poselow}_i}
\newcommand{\trackingpose}{\bm{\poselow}}
\newcommand{\rotlow}{\omega}
\newcommand{\rot}{\bm{\rotlow}}
\newcommand{\translow}{v}
\newcommand{\trans}{\bm{\translow}}
\newcommand{\hnorm}[1]{\left\lVert#1\right\rVert_{\gamma}}
\newcommand{\lnorm}[1]{\left\lVert#1\right\rVert}
\newcommand{\barate}{v_i}
\newcommand{\trackingrate}{v}
\newcommand{\imgpt}{\mathbf{u}_{i,k,j}}
\newcommand{\mappt}{\mathbf{X}_{j}}
\newcommand{\timet}[1]{\bar{t}_{#1}}
\newcommand{\mf}[1]{\text{MF}_{#1}}
\newcommand{\kmf}[1]{\text{KMF}_{#1}}
\newcommand{\Exp}{\text{Exp}}
\newcommand{\Log}{\text{Log}}
\newcommand{\mytextrm}[1]{\textrm{\tiny {#1}}}
\newcommand{\name}{{CASA~}} 

\newcommand{\todocite}[1]{\textcolor{blue}{Citation needed []}}

\makeatother

\newcommand{\shenlongsay}[1]{{\color{magenta}[#1]}}
\newcommand{\zeyuan}[1]{{\color{blue}[#1]}}
\newcommand{\yuefan}[1]{{\color{blue}{#1}}}


\maketitle

\begin{figure}[tbhp]
\centering
\vspace{-12mm}
\includegraphics[width=\linewidth]{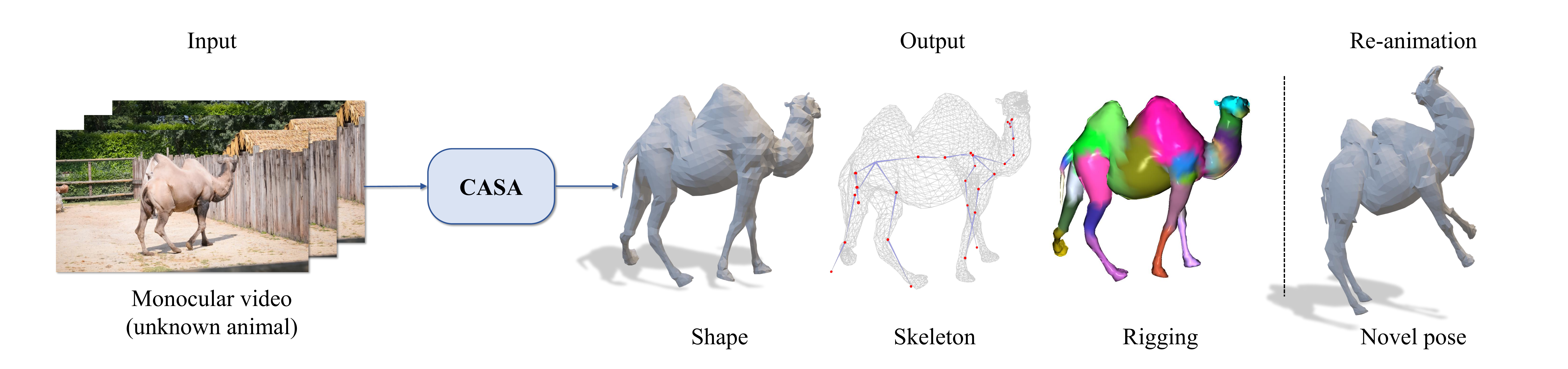}
\caption{Given a monocular video with an animal from unknown category, CASA jointly infers the articulated skeletal shape and rigging through optimization, which can be animated into novel poses.}
\label{fig:teaser}
\end{figure}

\begin{abstract}

Recovering the skeletal shape of an animal from a monocular video is a longstanding challenge. Prevailing animal reconstruction methods often adopt a control-point driven animation model and optimize bone transforms individually without considering skeletal topology, yielding unsatisfactory shape and articulation. 
In contrast, humans can easily infer the articulation structure of an unknown animal by associating it with a seen articulated character in their memory.  Inspired by this fact, we present \textbf{CASA}, a novel \textbf{C}ategory-\textbf{A}gnostic \textbf{S}keletal \textbf{A}nimal reconstruction method consisting of two major components: a video-to-shape retrieval process and a neural inverse graphics framework. During inference, CASA first retrieves an articulated shape from a 3D character assets bank so that the input video scores highly with the rendered image, according to a pretrained language-vision model. CASA then integrates the retrieved character into an inverse graphics framework and jointly infers the shape deformation, skeleton structure, and skinning weights through optimization. Experiments validate the efficacy of CASA regarding shape reconstruction and articulation. We further demonstrate that the resulting skeletal-animated characters can be used for re-animation. 

\end{abstract}
\section{Introduction}
\label{sec:intro}

Recovering the shape, articulation, and dynamics of animals from images and videos is a long-standing task in computer vision and graphics. 
Achieving this goal will enable numerous future applications for 3D modeling and reanimation of animals. 
Nevertheless, accurately reasoning about the geometry and kinematics of animals in the wild remains an ambitious problem for three reasons: 1) \textit{partial visibility} of the captured animal, 2) \textit{variability} of shape across different categories, and 3) \textit{ambiguity} of unknown kinematics. 
Take the camel in Fig.~\ref{fig:teaser} as an example -- to faithfully model it requires: 1) hallucinating the unobserved (\eg, occluded, back-side) region, 2) recovering its unique shape and scale, and 3) predicting its kinematic structure.

Remarkable progress has been made in addressing the aforementioned challenges by either exploiting richer sensor configurations (\eg, multi-view cameras~\cite{guo2019relightables,Joo_2017_TPAMI} or depth sensors~\cite{Newcombe2015DynamicFusionRA, OccupancyFlow, ren2021class}), or by making strong class-specific assumptions (\eg, humans~\cite{loper2015smpl, 2019PIFu,pifuhd,saito2021scanimate,yang2021s3,  ZhaoOplanes2022} or quadruped animals~\cite{Zuffi19Safari,zuffi2018lions,zuffi20173d}). However, these assumptions greatly limit the applicability of reconstruction systems, which fail to generalize to animals in the wild. 

We aim to reconstruct \emph{arbitrary articulated animals using monocular videos casually captured in the wild}. Recent works~\cite{2021LASR, 2021VISER, yang2021banmo, wu2021dove, park2021nerfies} demonstrate promising results. However, they often impose non-realistic assumptions on articulation, such as control point driven deformation~\cite{2021LASR, 2021VISER, yang2021banmo, wu2021dove} or free-form deformation~\cite{park2021nerfies, ren2021class}.
As a result, they fall short of the goal of modeling skeletal characters that can be realistically re-animated in downstream applications. Furthermore, there remains significant improvement space for the quality of the inferred animal shape. 

In this work, we propose \textbf{CASA}, a novel solution for \textbf{C}ategory-\textbf{A}gnostic \textbf{S}keletal \textbf{A}nimal reconstruction in the wild. \name jointly estimates an arbitrary animal's 3D shape, kinematic structure, rigging weight, and articulated poses of each frame from a monocular video (Fig.~\ref{fig:teaser}).
Unlike existing nonrigid reconstruction works~\cite{2021LASR, 2021VISER, wu2021dove}, we exploit a skeleton-based articulation model as well as forward kinematics, ensuring the realism of the resulting skeletal shape (\S~\ref{sec:articulation}). 
Specifically, we propose two novel components: a video-to-shape retrieval process and a {skeleton-based} neural inverse graphics framework. Given an input video, \name first finds a template shape from a 3D character assets bank so that the input video scores highly with the rendered image, according to a pretrained language-vision model~\cite{radford2021learning} (\S~\ref{sec:32}). 
Using the retrieved character as initialization, we jointly optimize bone length, joint angle, 3D shape, and blend skinning weight so that the final outputs are consistent with visual evidence, \ie, input video (\S~\ref{sec:33}). 

Another issue that hinders the study of animal reconstruction is that the existing datasets~\cite{bogo2017dynamic, hu2021sail, li20214dcomplete} lack realistic video footage and ground-truth labels across different dynamic animals. To address it, we introduce a photo-realistic synthetic dataset \textbf{\dataset{}}, which is generated using the physical-based rendering~\cite{visionblender} and rich simulated 3D assets~\cite{planetzoo}. \dataset{} is large-scale and consists of 251 different articulated animals. Importantly, \dataset{} provides ground-truth 3D shapes, skeletons, joint angles, and rigging, allowing evaluation of category-agnostic 4D reconstruction holistically (\S~\ref{sec:dataset}).  

We evaluate \name on both \dataset{} and the real-world dataset DAVIS~\cite{Perazzi2016davis}. Experiments demonstrate that \name recovers fine shape and realistic skeleton topology, handles a wide variety of animals, and adapts well to unseen categories. 
Additionally, we showed that \name reconstructs a skeletal-animatable character readily compatible with downstream re-animation and simulation tasks.



{In summary, we make the following contributions:
1) We propose a simple, effective, and generalizable video-to-shape retrieval algorithm based on a pretrained CLIP model.
2) We introduce a novel neural inverse graphics optimization framework that incorporates stretchable skeletal models for category-agnostic articulated shape reconstruction.
3) We present a large-scale yet diverse skeletal shape dataset PlanetZoo.
}

    



\section{Related work}

\begin{figure}
\centering
\vspace{-3mm}
\includegraphics[width=\linewidth]{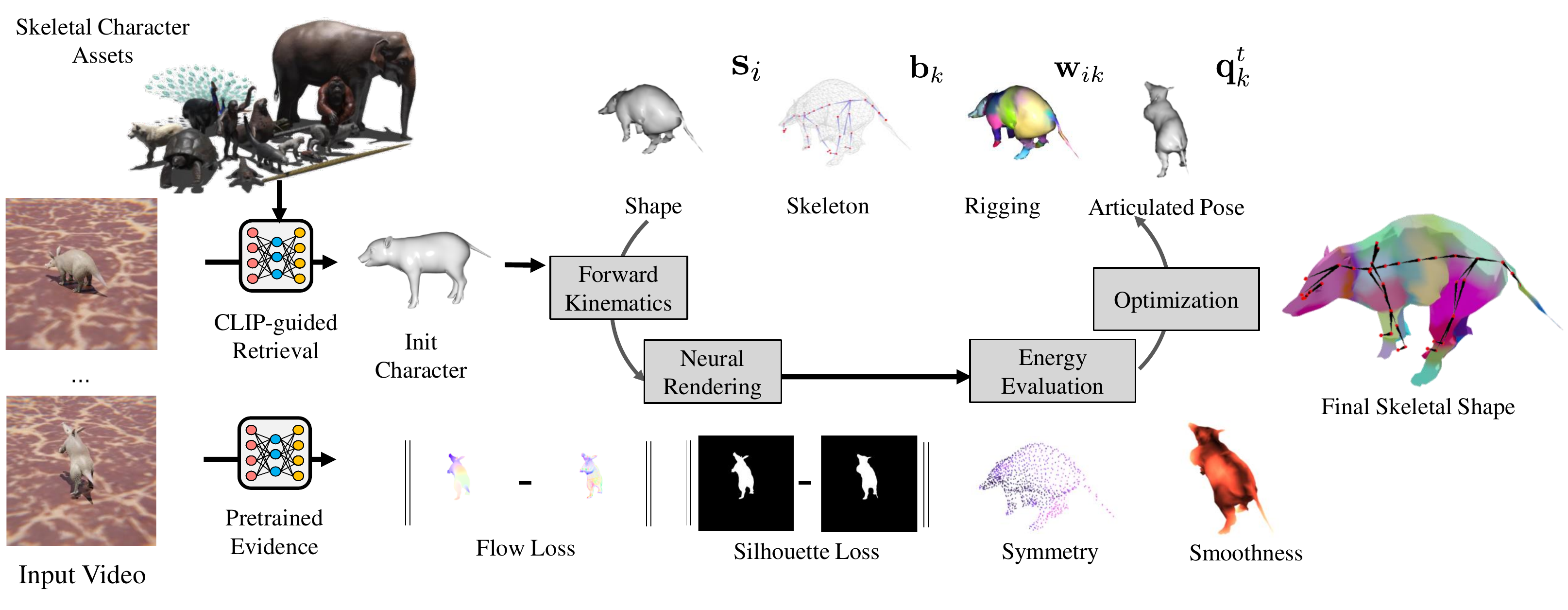}
\caption{\textbf{Overview.} Given an input video, A video-to-shape retrieval process is first conducted with the guidance of pre-trained CLIP (\S~\ref{sec:32}). Initialized by the retrieved shape, we jointly optimize shape, skeleton (bone length and joint angle), and skinning through inverse rendering (\S~\ref{sec:33}).}
\label{fig:overview}
\end{figure}

\textbf{Category-specific nonrigid reconstruction.}
Parametric model (template) is a classic approach for category-specific nonrigid reconstruction\cite{LEAP,kanazawa2018endhmr,biggs2020left,yang2021s3,yang2020recovering,ma2022virtual,liu2021semi,ruegg2022barc} . Templates are built for various objects including human body~\cite{loper2015smpl,anguelov2005scape}, hand~\cite{romero2022embodied}, face~\cite{li2017learning, blanz1999morphable}, and animals~\cite{zuffi20173d,biggs2020left}. Template-free methods~\cite{ 2018CMR, li2020online, kulkarni2020articulation, tulsiani2020implicit, saito2021scanimate, wu2021dove, kokkinos2021learning,palafox2021npms,weng_humannerf_2022_cvpr} study the problem of predicting nonrigid objects directly for a specific category. However, these methods heavily rely on strong category priors such as key-points annotations~\cite{2018CMR}, canonical shapes~\cite{ucmr_goel2020shape}, temporal consistency constraints~\cite{li2020online}, or canonical surface mapping~\cite{kokkinos2021learning}, making them hard to generalize to category-agnostic setting. 


\textbf{Category-agnostic nonrigid reconstruction.}
Classic approaches~\cite{bregler2000recovering,kong2019deep,sidhu2020neural,noguchi2022watch} utilize Non-Rigid Structure from Motion (NRSfM) for non-rigid reconstruction, which requires point correspondences and strong priors on shape or motion. 
{However, NRSfM does not work well for monocular videos in the wild, in which obtaining long-range correspondences is hard.}
Recent works adopt learning-based approaches via either dynamic reconstruction~\cite{lei2022cadex, ren2021class} or inverse graphics optimization~\cite{2021LASR, 2021VISER,yang2021banmo}. 
However, these methods do not reason skeletal kinematics and suffer from incorrect shape prediction when certain parts are occluded or invisible.

\textbf{Animatable shape.}
Humans and animals undergo skeleton-driven deformation. A skeletal shape often consists of two parts: a surface mesh representation used to draw the character and a skeleton structure to control the motion. Each skeleton is a hierarchical set of bones. Each bone has associated with a portion of vertices (skinning). The bone's transformation is determined through a forward kinematics process. As the character is animated, the bones change their transformation over time. 
Linear Blend Skinning (LBS)~\cite{magnenat1988joint,lewis2000pose} is a standard way for modeling skeletal deformation, which deforms each vertex of the shape based on a linear combination of bone transformations. For improvement, multi-weight enveloping~\cite{wang2002multi,merry2006animation} is used to overcome the issue of shape collapse near joints when related bones are rotated or moved. Dual-quaternion blending (DQB)~\cite{kavan2008geometric} adopts quaternion rotation representation to solve the artifacts in blending bone rotations, and STBS~\cite{2011stretch} extends LBS to include the stretching and twisting of bones.
Recent works~\cite{2021LASR,2021VISER,yang2021banmo} employs LBS for modeling the motion of shapes recovered from video clips. 
However, these methods do not enforce a skeleton-based forward kinematic structure. Hence their recovered animated shapes are not interpretable and cannot be directly used in skeletal animation and simulation pipeline. 

\textbf{Language-vision approaches.}
Self-supervised language-vision models have gone through rapid advances in recent years~\cite{sun2019videobert,wang2021clip,ramesh2022hierarchical} due to their impressive generalizability. 
The seminal work CLIP~\cite{wang2021clip} learns a joint language-vision embedding using more than 400 billion text-image pairs. The learned representation is semantically meaningful and expressive, thus has been adapted to various downstream tasks~\cite{zhou2022learning,xu2021videoclip,rao2022denseclip,zhang2022pointclip,vinker2022clipasso}. In this work, we adopt CLIP in the retrieval process.

\textbf{3D reconstruction dataset.}
A plethora of synthetic 3D datasets~\cite{replica19arxiv,zheng2020structured3d,johnson2017clevr,roberts2021hypersim} and interactive simulated 3D environments~\cite{savva2019habitat,shen2020igibson,dosovitskiy2017carla, wang2020tartanair} have been proposed in recent years. ShapeNet~\cite{chang2015shapenet} provides a benchmark for common static 3D object shape reconstruction. Among all of these datasets, only a few aim for dynamic object reconstruction~\cite{li20214dcomplete}. Given that the synthetic dataset has a large domain gap to  real-world settings, photo-realistic rendered samples are strongly preferred. To this end, we propose a photo-realistic synthetic dataset PlanetZoo to study the dynamic animal reconstruction problem. PlanetZoo contains high-fidelity assets and covers a wide range of animal categories. 
\section{Category-agnostic skeletal animal reconstruction}
\name aims to reconstruct various dynamic animals in-the-wild from monocular videos. It takes as input the RGB frames $\{\bI_t\}_{1...T}$ from a monocular video, object masks~$\{\bM_t\}_{1...T}$, and optical flow maps~$\{\bF_t\}_{1...T}$ computed from consecutive frames. From these inputs, we aim to recover an animal shape $\bs_0$ and its articulated deformed shape $\bs_t$ at each time $t$. 

Fig.~\ref{fig:overview} demonstrates an overview of our approach. Our method exploits the skeletal articulation model and forward kinematics to ensure the realism of the resulting skeletal shape (\S~\ref{sec:articulation}). Our method consists of two phases. In the retrieval phase (\S~\ref{sec:32}), CASA finds a template character from an existing asset bank based on the similarity to the input video in the deep embedding space using a pre-trained encoder. The retrieved template is fed into the neural inverse graphics phase as initialization(\S~\ref{sec:33}). Finally, we jointly reason the final shape, skeleton, rigging and the articulated pose of each frame through an energy minimization framework.

\begin{figure}[t]
\centering
\includegraphics[width=0.85\linewidth]{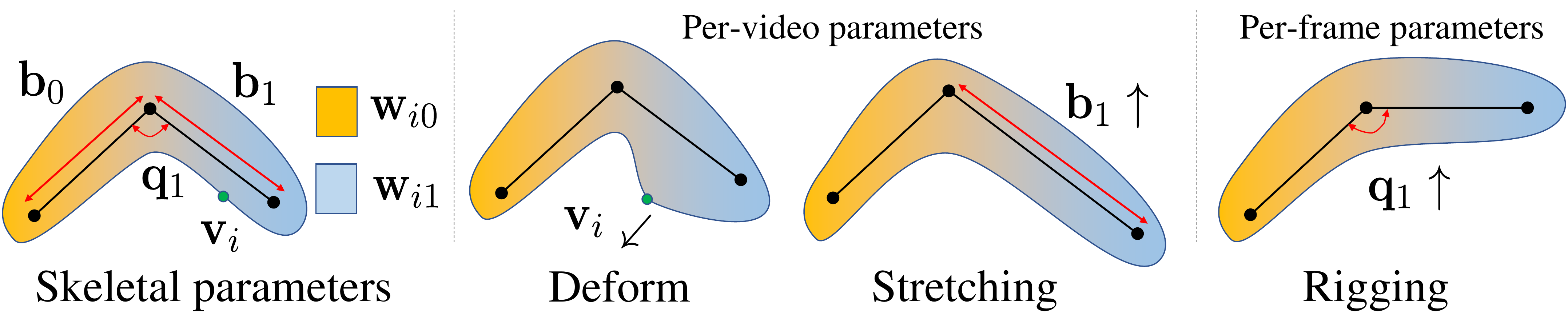}
\caption{\textbf{Skeletal shape parametric model.} Our parametric model consists of joint angles $\bq$, bone length $\bb$, skinning weight $\bw$, and vertex positions $\bv$. Joint angles $\bq$ change per-frame and the others are universal. 
Utilizing this model, animal shapes are tuned by vertex deformation as well as the stretching of bones. The target articulation would be fit by predicting per-frame joint angles.}
\label{fig:articulation}
\end{figure}

\subsection{Articulation model}
\label{sec:articulation}

\noindent\textbf{Skeletal parameters.} 
We exploit a stretchable bone-based skeletal model for our deformable shape parameterization ,as shown in Fig.~\ref{fig:articulation}. This articulated model consists of three components: 1) a triangular mesh consisting of a set of vertices and shapes in the canonical pose, describing the object's shape; 2) a set of bones connected by a kinematic tree. Each one has a bone length parameter and associated rigging weight over each vertex; 3) a joint angle describing the relative transformation between each adjacent bone. To summarize, a deformed shape $\bs_t$ at time $t$ can be represented as:
\begin{equation}
\label{eq:params}
\mathbf{s}_t = \left\{ \left\{\mathbf{v}_i \in \bbR^3\right\}_N, \left\{\mathbf{f}_j \in \mathbb{I}^3\right\}_T, \left\{\mathbf{w}_{i} \in \Delta^K\right\}_N, \left\{b_k \in \bbR^1\right\}_K, \left\{\mathbf{q}_k^t \in \SO3\right\}_{K}\right\}
\end{equation}
where $\mathbf{v}_i$ is the vertex position, $\mathbf{f}_j$ represents a triangle face parameterized as a tuple of three vertex indices, $\mathbf{w}_i$ is the rigging weight constrained in a K-dimensional simplex:
$
\Delta^K = \left\{(w_0,\dots,w_K)\in\mathbb{R}^K ~|~ \sum_{k = 0}^K w_k = 1 \textrm{ and } w_k \ge 0 \right\}$; 
$b_k$ is the scalar bone length for each bone and $\mathbf{q}_k^t$ is the joint angle for each bone at a time $t$, represented as a unit quaternion in the $\SO3$ space. Note that all the variables except joint angles are shared across time.

Our skeletal model is grounded by the nature of many articulated objects. 
Unlike the commonly used control-point based articulation~\cite{Newcombe2015DynamicFusionRA, 2021VISER, 2021LASR}, our model reflects the constraints imposed by bones and joints, leading to more natural articulated deformation. Compared to category-specific parametric models~\cite{loper2015smpl, zuffi20173d}, it is more flexible and generalizable. 

\noindent\textbf{Forward kinematics.}
We use the forward kinematic model~\cite{lavalle2006planning} to compute the transformation of each bone along the kinematic tree. Specifically, the rigid transformation of a bone is uniquely defined by the joint angles and bone length scales of the bone itself and its ancestors along the kinematic tree. Given a bone $k$, the rigid transform between its own frame and the root can be computed by recursively applying the relative rigid transformation along the chain:
\begin{equation}
     \mathbf{T}_k^t(\mathbf{q}^t_\mytextrm{\tiny ans(k)}, \mathbf{b}_\mytextrm{\tiny ans(k)}) = \mathbf{T}_{\mytextrm{\tiny  pa}(k)}^t \mathbf{T}^t_{k, \mytextrm{pa}(k)}(\mathbf{q}_k^t, b_k) = \prod_{k\prime \in \mytextrm{ans}(k)} \mathbf{T}^t_{k^\prime, \mytextrm{pa}(k^\prime)}(\mathbf{q}_{k^\prime}^t, b_{k^\prime})
\label{eq:FK}
\end{equation}
where $\mathbf{T}_k^t$,$\mathbf{T}_{\mytextrm{\tiny pa}(k)}^t$ is the transformation of bone $k$ and its parent node at frame $t$. $\mathbf{q}_\mytextrm{\tiny ans(k)}, \mathbf{b}_\mytextrm{\tiny ans(k)}$ are the joint angles and bone lengths from all the ancestors of target bone $k$. $\mathbf{T}^t_{k, \mytextrm{pa}(k)}(\mathbf{q}_k^t, b_k)$ is the relative rigid transform between the bone and its parent frame, consists of a joint center translation along z axis decided by the bone length $b_k \mathbf{e}_z = [0, 0, b_k]$ and a rotation around the joint center $\bR(\bq^t_k)$.

\noindent\textbf{Linear Blend Skinning (LBS).} Given the rigid transformation of each bone, we adopt LBS to compute the transform of each point. Specifically, the deformation of each point $\mathbf{v}$ is decided by a linear composing the rigid transforms of bones through its skinning weight:
$
    \mathbf{v}_i^t= \sum_{k = 0}^K w_{i,k} \mathbf{T}_k^t \mathbf{v}_i
$
where $w_{i,k}$ is the rigging weight. $\mathbf{T}_k^t$ is the transformation of bone $k$ at frame $t$ as defined in Eq.~\ref{eq:FK}. For simplicity, we omit its dependent variables.

\noindent\textbf{Stretchable bones.} 
The aforementioned nonrigid deformation assumes independent shape and bone lengths. 
Naturally, the shape should deform accordingly as bone length changes.  
To model such a relationship, we exploit a stretchable bone deformation model~\cite{2011stretch}. For each bone, we compute an additional scaling transform along the bone vector direction based on its bone length change and apply this scaling transform to any points associated with this bone. Fig.~\ref{fig:articulation} provides an illustration of stretching. For mathematical details on stretching, we refer readers to our supplementary document.  

Stretchable bones bring a two-fold advantage: 1) it allows us to model shape variations within the same topological structures; 2) it makes the topological structure adjustable by ``shrinking'' the bones, e.g., a quadrupedal animal can be evolved into a seal-like skeletal model through optimization. 

\noindent\textbf{Reparameterization.} 
Optimizing each vertex position offers flexibility yet might lead to undesirable mesh due to the lack of regularization. We use a neural displacement field to re-parameterize the vertex deformation to incorporate regularization, such as smooth and structured vertex deformation. We use a coordinate-based multi-layer perceptron (MLP) to define this displacement field $\mathcal{V}$. In addition, we incorporate a global scale scalar $\mathbf{u}$ in our framework to handle the shape misalignment between our initialization and the target. The position of each vertex is defined as:
\begin{equation}
    \mathbf{v}_i' = u\mathbf{v}_i + \mathcal{V}_\theta(u\mathbf{v}_i)
\end{equation}
where $u\mathbf{v}_i$ is the scaled position at the canonical pose, $\mathcal{V}_\theta(u\mathbf{p}_i)$ represents the vertex offset parameterized by $\theta$, and $\mathbf{v}_i'$ is the updated position for vertex $i$. During inference, instead of directly optimizing $\bv_i$, we will only optimize the parameters of the displacement network. This displacement field reparameterization allow us to smoothly deform canonical shape during inference. In practice we find this implicit regularization compares favorably over explicit smoothness terms such as as-rigid-as-possible and laplacian smoothness.

\subsection{Video-to-shape retrieval}
\label{sec:32}
Directly optimizing all skeletal parameters defined in~\equref{eq:params} is challenging due to the variability and highly structured skeletal parameterization. Many deformable objects share similar topology structures, albeit with significant shape differences. To address it, we propose initializing skeletal shapes in a data-driven manner by choosing the best matching template from an asset bank. 

Videos and skeletal shapes are in two different modalities; hence establishing a similarity measure is hard. 
Inspired by the recent success of language-vision pretraining models for 3D~\cite{wang2021clip, text2mesh}, we utilize realistic rendering and pretraining image embedding models~\cite{radford2021learning} to bridge this gap. Specifically, we first pre-render video footage for each character in the asset bank through a physically based renderer~\cite{visionblender}. The environment lighting, background, articulated poses, and camera pose for each video are randomized to gain diversity and robustness.   We then extract the image embedding features for each video using CLIP~\cite{radford2021learning}, which is a language-vision embedding model that is pretrained on a large-scale image-caption dataset. It captures the underlying semantic similarity between images well despite the large appearance and viewpoint difference. This is particularly suitable for retrieving shapes with similar kinematic structures, as the kinematics of animals is often related to semantics. 

During inference, we extract embedding features from a given input video and measure the {L2 distance} between the input video and the rendered video of each object in the embedding space. We then select the highest-scoring articulated shape as the retrieved character. {Please refer to the 
supplementary material for implementation details. }


\subsection{Neural inverse graphics via energy minimization}
\label{sec:33}

We expect our final output skeletal shape to 1) be consistent with the input video observation; 2) reflect prior knowledge about articulated shapes ,such as symmetry and smoothness. Inspired by this, we formulate ours as an energy minimization problem. 

\noindent\textbf{Energy formulation.}  We exploit visual cues from each frame including segmentation mask $\{\bM_t\}$ and optical flow $\{\bF_t\}$ through off-the-shelf neural networks~\cite{teed2020raft,2021LASR}. We also incorporate two types of prior knowledge, including the motion smoothness for joint angles $\mathbf{q}_k^t$ of bones across frames, as well as the symmetry constraint for the per-vertex offset $\mathbf{v}_i$ at the reference frame.
In particular, let $\mathcal{I}=(\{\bI_t\},\{{\bM}_t\},\{{\bF}_t\})$ be the input video frames and corresponding visual cues, and $\{\mathbf{s}_t\}$ be the predicted shapes of all frames, we formulate the energy for minimization as:
\begin{equation}
    \min\limits_{\{\mathbf{s}_t\}_{1...T}}  \lambda_1 E_\mytextrm{\tiny cue}(\{\mathbf{s}_t\},\mathcal{I}) + \lambda_2 E_\mytextrm{smooth}({\{\mathbf{s}_t\}})+\lambda_3 E_\mytextrm{symm}(\{\mathbf{s}_t\})
\label{eq:energy}
\end{equation}
where $E_\mytextrm{\tiny cue}$ measures the consistency between the articulated shape and the visual cues; $E_\mytextrm{smooth}$ promotes smooth transitions overtime; $E_\mytextrm{symm}$ encodes the fact that many articulated objects are symmetric.  The three energy terms complement each other, helping our model capture motions of the object according to visual observations, as well as constraining the deformed object to be natural and temporally consistent. We describe details of each term below.

\noindent\textbf{Visual cue consistency.} The visual cue consistency energy measures the agreement between the rendered maps of the articulated shape and 2D evidence (flow and silhouettes) from videos. We use a differentiable renderer~\cite{liu2019soft} to generate projected object silhouettes $\{\mathbf{M}_t\}$. Additionally, we project vertices of predicted shapes at two consecutive frames to the camera view and compute the projected 2D displacement to render optical flow, following the prior work~\cite{2021LASR}. We leverage PointRend~\cite{2020Pointrend} for object segmentation and volumetric correspondence net~\cite{2019VCN} for flow prediction. The energy measures the difference between the rendered and inferred cues in $\ell_2$ distance: 
\begin{equation}
    E_\mytextrm{cue} = \sum_t\left\{ \| \bM_t - \pi_\mytextrm{seg}(\mathbf{s}_t)\|^2 + \beta \| \bF_t - \pi_\mytextrm{flow}(\mathbf{s}_t)\|^2 \right\}
\label{eq:cue}
\end{equation}
where $\beta$ is the trade-off weight; $\pi_\mytextrm{flow}(\mathbf{s}_t)$ is the rendered the 2D flow map for each pixel given the deformed shape $\mathbf{s}_t$; $\pi_\mytextrm{seg}(\mathbf{s}_t)$ is the rendered object mask. Similar to previous inverse rendering work~\cite{2021VISER}, the object-camera transform is encoded as the root node transform. 

\noindent\textbf{Motion smoothness.} This energy term encodes that motions of animals should be smooth and continuous across frames. We impose a constraint to ensure that there is little difference in joint angles of one bone from two consecutive frames should be slight. We implement this by computing the multiplication of the joint quaternion at the current frame and the inverse of the joint quaternion at the next frame, which should be close to an identity quaternion $\mathbf{q} = (0,0,0,1)$:
\begin{equation}
    E_\mytextrm{smooth} = \sum_t \sum_k \| (\mathbf{q}_k^t)^{-1} \circ \mathbf{q}_k^{t + 1} - \mathbf{q}\|^2
\label{eq:smooth}
\end{equation}
where $\circ$ is the quaternion composition operator.

\textbf{Symmetry offset.} This term encourages the resulting shape in canonical space to be symmetric at the reference frame. It is inspired by the fact that most animals in the real world are symmetric if putting them into a certain canonical pose (e.g., `T-pose` for bipedal animals). 
{Following previous works~\cite{2021LASR},} we enforce this property at the canonical shape (where the joint angles are all zero). Specifically, we calculate the chamfer distance for measuring the similarity between the set of vertices $\{\mathbf{v}_i\}$ under the canonical shape and its reflection:
\begin{equation}
    E_\mytextrm{symm} = \mathcal{L}_\mytextrm{cham}((\{\mathbf{v}_i\}),\mathbf{H}(\{\mathbf{v}_i\}))
\label{eq:symm}
\end{equation}
where $\mathbf{H}$ is the Householder reflection matrix.

\subsection{Inference}
We reason the skeletal shape by minimizing the energy defined in Eq.~\ref{eq:energy}. Our optimization variables include the vertex position $\mathbf{v}_i$ at the canonical pose, the rigging weight $\mathbf{w}_i$, bone length $\mathbf{b}_k$ as well as the joint angles $\mathbf{q}_k^t$. All the variables except joint angles are shared across time, and joint angles are optimized per frame. 

\textbf{Initialization.} We initialize the vertex position, bone length, and the rigging weight $\mathbf{v}_i, \mathbf{b}_k, \mathbf{w}_i$ of canonical shape using the retrieved template character. The skeleton tree structure of this character is also taken as the basis of our skeletal parameterization. We initialize all the joint rotations as a unit quaternion. 


\textbf{Optimization.} The energy function is fully differentiable and can be optimized end-to-end. We use Adam optimizer~\cite{2014ADAM} to learn all the optimization variables. We adopt a scheduling strategy to avoid getting stuck at a local minimum. Specifically, we first optimize the mesh scaling factor based on silhouette correspondences. We then jointly update the bone length scale, joint angles, and the neural offset field by minimizing the energy function defined in Eq.~\ref{eq:energy}.

\begin{table}[!t]
\small
\centering
\vspace{-3mm}
\caption{Quantitative results on \dataset{}.} 
\label{tab:mesh}
\begin{tabular}{rccccc}
\toprule
Method  & mIoU $\uparrow$ & mCham $\downarrow$ & Skinning $\downarrow$ & Joint $\downarrow$ & Re-animation $\downarrow$\\
\cmidrule(l){1-6}
ViSER~\cite{2021VISER} & 0.190 & 0.133 & 6.398 & 0.416 & 0.028 \\
LASR~\cite{2021LASR} & 0.306 & 0.094 & \textbf{2.576} & 0.158 & 0.033 \\
CASA (ours) & \textbf{0.512} & \textbf{0.053} & 3.288 & \textbf{0.089} & \textbf{0.010} \\
\bottomrule
\end{tabular}
\end{table}

\begin{figure*}
\vspace{-3mm}
    \centering
    \includegraphics[width=.9\linewidth]{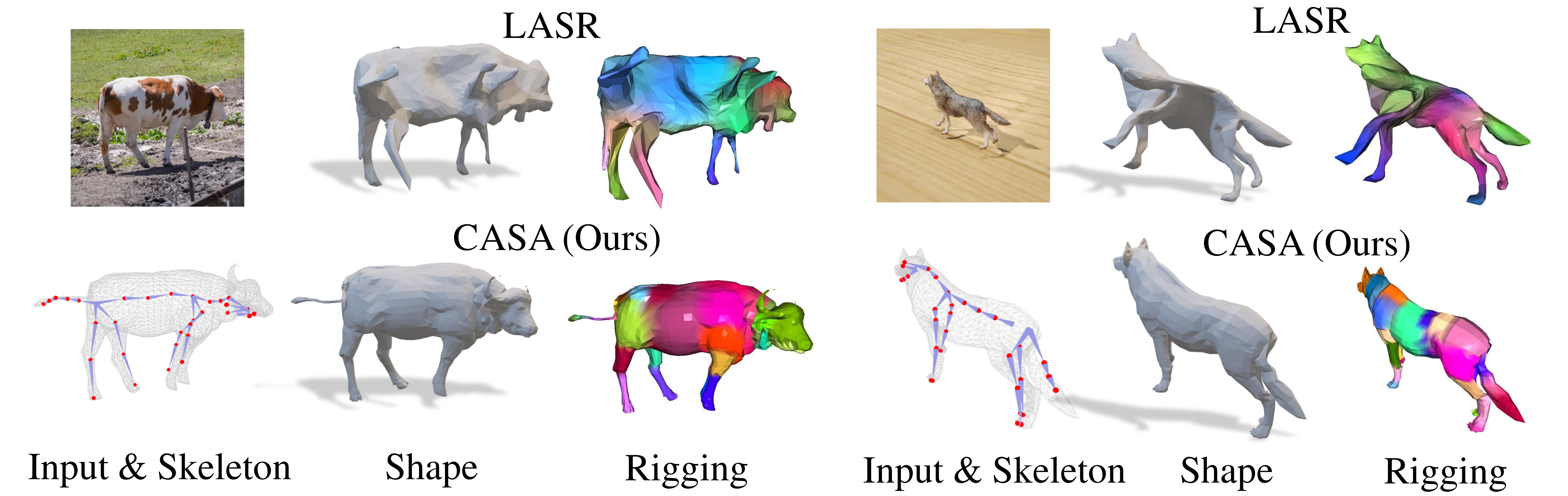}
    \vspace{-1mm}
    \caption{\textbf{Reconstructed articulated shape.} Top-left: input; top: LASR; bottom: ours.}
    \label{fig:qual}
    \vspace{-5mm}
\end{figure*}





\section{\dataset{} dataset}
\label{sec:dataset}

Benchmarking 4D animal reconstruction requires a large amount of nonrigid action sequences with ground truth articulated 3D shapes. Towards this goal, we construct a synthetic dataset \dataset{} consisting of 
hundreds of animated animals with textures and skeletons from different categories.
Appendix Fig.~\ref{fig:dataset} depicts a snapshot of the assets and rendering images from our dataset, demonstrating the diversity and quality. 

\noindent\textbf{Data generation.}
We extracted animal meshes from the zoo simulator Planetzoo~\cite{planetzoo}. Cobra-tools~\cite{Cobra_tools} are used to extract those meshes along with their skeleton. With the extracted mesh and skeleton, we further render RGB maps, segmentation masks and optical flow for each frame using Blender.

\noindent\textbf{Assets.} To set a diverse environment, we set the background with random HDRI pictures for environmental lighting, and set floor textures with random materials from ambientCG\footnote{{https://ambientcg.com/}}. To reduce the gap between synthetic and real, we  set the location of the light source along with its strength to simulate different environments in real-world situations. 

\noindent\textbf{Camera.} To generate realistic action sequences, we randomly change camera locations between every 12 frames, resulting in a constant view-point change following the animal. The camera is allowed to rotate at a certain angle, ranging from $15^{\circ}$ to $22.5^{\circ}$. 

\noindent\textbf{Articulation.} In order to obtain animated animal sequences, for every 12 frames, 8 bones are selected from the skeleton tree with a transformation attached. The angle value of rotation for each bone is sampled from a uniform distribution. By doing this, we are able to cover the whole action space, providing diverse action sequences.

\noindent\textbf{Rendering.} We generate silhouettes, optical flow, depth map, camera parameters, and RGB images using Vision Blender~\cite{visionblender}. The physically based renderer is capable of showing fine details such as fur, making the rendering results more realistic. For each animal in the dataset, 180 frames are rendered.

\begin{figure*}
\centering
\includegraphics[width=\linewidth]{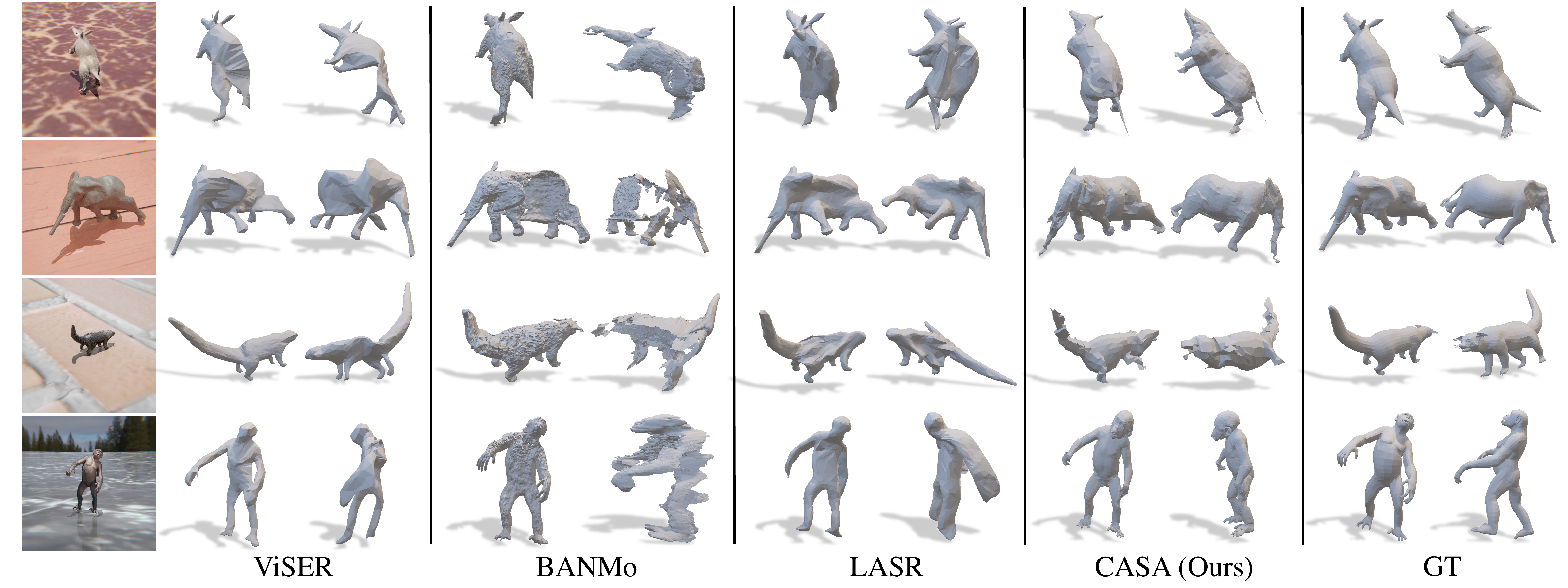} 
\caption{\textbf{Qualitative comparison of our method on \dataset{}.}} 
\label{fig:sync}
\includegraphics[width=\linewidth]{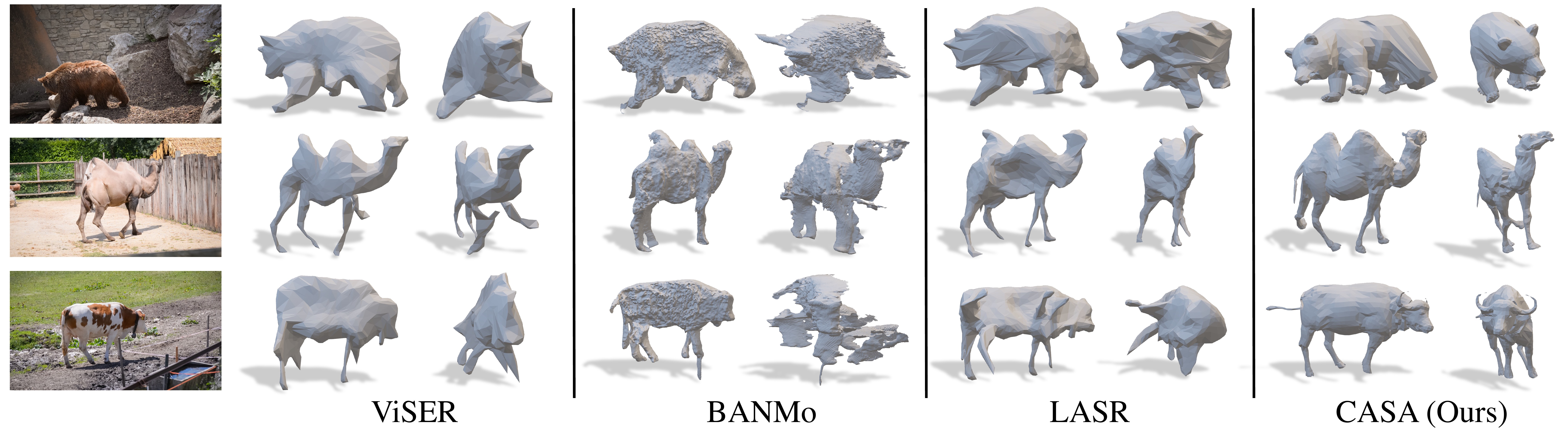}
\caption{\textbf{Qualitative comparison of our method on real-world videos from DAVIS.}} 
\label{fig:real}
\vspace{-7mm}
\end{figure*}


\section{Experiments}

In this section, we first introduce our experimental setting (\S~\ref{sec:setup}). We then compare \name{} against a comprehensive set of articulated baselines in various reconstruction and articulation metrics on both simulation~(\S~\ref{sec:simulation}) and real-world datasets~(\S~\ref{sec:realworld}). 
Finally, we demonstrate our inferred shape can be used for downstream reanimation tasks~(\S~\ref{sec:reanimation}). 

\subsection{Experimental setup}
\label{sec:setup}

\noindent\textbf{Benchmarks.} We validate our proposed method on two datasets. Our proposed photorealistic rendering dataset \dataset{} as well as a real-world animal video dataset {DAVIS}\cite{Perazzi2016davis}, containing multiple real animal videos with mask annotations. For \dataset{}, we choose 24 out of 249 total animals for testing and use the rest for validation and training. The testing dataset includes diverse articulated animals from unseen categories, including multiple quadrupeds, bipeds, birds categories, as well as unseen articulated topology such as pinnipeds.\\ 
\noindent\textbf{Metrics.} 
We measure the reconstruction quality by Intersection over Union (IOU) and Chamfer distance, as well as skinning distance, joint distance and re-animation quality on \dataset{}. 

1) \emph{mean Intersection Over Union(mIOU)} measures the volumetric similarity between two shapes. We voxelized the reference and the predicted shape into occupancy grids and calculate the IOU ratio. 

2) \emph{mean Chamfer Distance(mCham)} computes the bidirectional vertex-to-vertex distances between the reference  and the predicted shape.

3) \emph{Joint} is the symmetric Chamfer Distance between joints. We evaluate CD-J2J following~\cite{RigNet}. Given a predicted shape, we compute the Euclidean distance between each joint and its nearest joint in the reference shape, then divide it by the total joint number.
    
4) \emph{Skinning} distance measures the similarity between the skinning weights. We first extract vertices associated with each joint using skinning weight. For each pair of joints from GT and prediction, we calculate the Chamfer distance between their associated vertices. Finally, we exploit the Jonker-Volgenant algorithm to find the minimum distance matching between prediction and reference.
    
5) \emph{Re-animation} measures how well we can re-pose an articulated shape to a target shape. Specifically, we minimize the Chamfer Distance between the reference shape and the predicted shape by optimizing the joint angles of each bone for skeletal shape, or the rigid transform for control point-based shape. We consider this a holistic metric that jointly reflects the quality of shapes, skinning, and skeleton.

\noindent\textbf{Baselines.} We compare with state-of-the-art approaches for monocular-video articulated shape reconstruction ,including  LASR~\cite{2021LASR}, ViSER~\cite{2021VISER}, and BANMo~\cite{yang2021banmo}\footnote{We thank Gengshan Yang for sharing the code and providing guidance in running the baseline algorithms.}. Similar to our method, they utilize 2D supervision of videos for training including segmentation masks and optical flows.  We download the open-source version of them from GitHub. For the input data, we use our ground truth silhouette either in \dataset{} dataset or DAVIS\cite{Perazzi2016davis} dataset. We follow the optimization scripts in their code and get the baseline results.

3D reconstruction from a monocular video inevitably brings scale ambiguities. To alleviate the issue of scale differences, for every predicted shape among all the competing baselines, we conduct a line search to find the optimal scale that maximizes the IOU between the reference and the predicted shape.

\subsection{Results on \dataset{}}
\label{sec:simulation}

We present quantitative results on \dataset{} in Tab. \ref{tab:mesh}. CASA achieves higher performance on metrics mIOU and mCham, demonstrating a shape with higher fidelity. Our approach also outperforms other competing methods in reanimation, demonstrating the superior performance of holistic articulated shape reasoning and the potential for downstream reanimation tasks. CASA compares less favorably to LASR in the skinning quality. We conjecture this is due to additional degrees of freedom benefits brought by joint control point articulation. 

Fig.~\ref{fig:qual} and Fig.~\ref{fig:sync} depict a few qualitative examples of all competing algorithms on \dataset. We show the reconstructed mesh from both the camera view and the opposite view. 
The figure shows that CASA can recover accurate shape across various view angles under partial observation. In contrast, baselines fail to reconstruct reliable 3D mesh when the objects are partially observed. In addition, CASA also produces meshes with higher fidelity and local details in the visible region.

\subsection{Real-world reconstruction}
\label{sec:realworld}
We demonstrate qualitative results of the competing algorithms on the real-world animal video dataset {DAVIS}~\cite{Perazzi2016davis} in Fig \ref{fig:real} and Fig.~\ref{fig:qual}. This figure shows that both our method and baselines can get good reconstruction results from the camera perspective. But both two baselines fail to reconstruct those unseen parts. In contrast, CASA can reliably reason the shape and articulation in the unseen regions, thanks to its symmetry constraints, skeletal parameterization, and 3D template retrieval.


\begin{figure*}
\vspace{-5mm}
    \centering
    \includegraphics[width=\linewidth]{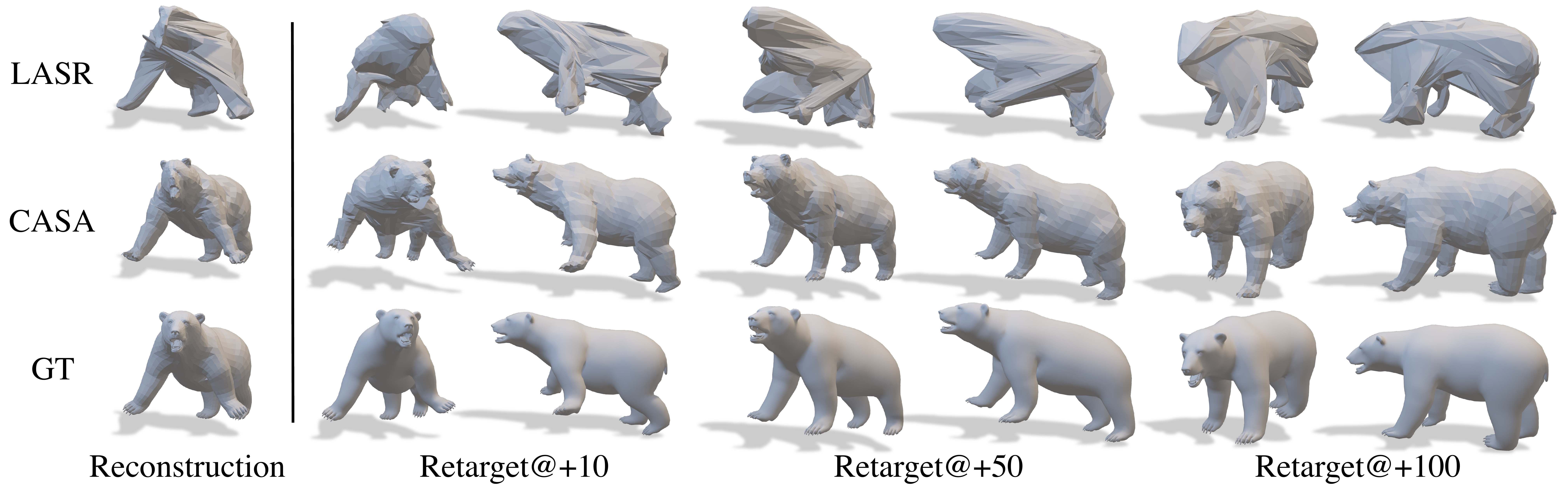}
    \caption{\textbf{Reanimation results.}}
    \label{fig:reanimation}
\vspace{-5mm}
\end{figure*}

\subsection{Reanimation}
\label{sec:reanimation}

We now show how the reconstructed articulated 3D shape can be retargeted to new poses. Given an inferred skeletal shape and the target GT shape at a different articulated pose, we apply an inverse kinematic to compute the articulated pose to reanimate the inferred shape. Specifically, we optimize the articulated transforms such that the reanimated mesh is as close to the target in Chamfer distance. Joint quaternions are optimized for skeletal mesh, and the articulation transforms are optimized for the control-point-based method~\cite{2021LASR}. Fig.~\ref{fig:reanimation} shows a comparison between the retargeted meshes of our methods and LASR on \dataset{}, with a GT target mesh as a reference. We observe that our retargeted meshes look realistic and accurate and preserve geometric details. Despite having more degrees of freedom in reanimation, LASR fails to produce realistic retarget results. Tab.~\ref{tab:mesh} reports a quantitative comparison in chamfer distance between the retargeted and the reference mesh, demonstrating our approach outperforms competing methods by a large margin.

\begin{table}[h]
    \parbox[t]{.48\linewidth}{
        \centering
        \caption[]{\centering Ablation study on energy terms\footnotemark}
        \label{tab:abla-energy}
        \resizebox{6.6cm}{!}{
        \begin{tabular}{lcccc}
        \toprule
        Method  & mIOU $\uparrow$ & mCham $\downarrow$ & Skinning $\downarrow$ & Joint $\downarrow$ \\
        \cmidrule(l){1-5}
        CASA         & \textbf{0.512} & 0.053 & 3.288 & \textbf{0.089} \\
        w/o $E_\mytextrm{mask}$ & 0.011 &  *  &  * & * \\
        w/o $E_\mytextrm{flow}$ & 0.427 & 0.350 & 3.290 & 0.636 \\
        w/o $E_\mytextrm{symm}$ & 0.387 & *  & 4.008 & * \\
        w/o $E_\mytextrm{smooth}$ & 0.449 & \textbf{0.041} & \textbf{3.228} & 0.330 \\
        \bottomrule
        \end{tabular}
        }
    }
    \hfill
    \parbox[t]{.48\linewidth}{
        \centering
        \caption{\centering Ablation study on optimization settings.} 
        \label{tab:abla-opti}
        \resizebox{7cm}{!}{
        \begin{tabular}{lcccc}
        \toprule
        Method  & mIOU $\uparrow$ & mCham $\downarrow$ & Skinning $\downarrow$ & Joint $\downarrow$ \\
        \cmidrule(l){1-5}
        CASA & \textbf{0.512} & \textbf{0.053} & 3.288 & \textbf{0.089} \\
        w/o offset & 0.144 & 0.343 & \textbf{3.174} & 0.783 \\
        w/o disp.~field & 0.235 & 0.055 & 3.218 & 0.262 \\
        w/o skeleton & 0.372 & 0.060 & 3.245 & N/A \\
        w/o scaling & 0.299 & 0.072 & 3.329 & 0.356 \\
        \bottomrule
        \end{tabular}
        }
    }
\end{table}
\footnotetext{*: extremely large value}
\begin{table}[h]
    \parbox[t]{.48\linewidth}{
    
        \centering
        \caption{\centering Ablation study on our retrieval strategy.}
        \label{tab:abla-retri}
        \resizebox{7cm}{!}{
        \begin{tabular}{rcccc}
        \toprule
        Method  & mIOU $\uparrow$ & mCham $\downarrow$ & Skinning $\downarrow$ & Joint $\downarrow$ \\
        \cmidrule(l){1-5}
        CASA & \textbf{0.512} & \textbf{0.053} & \textbf{3.288} & \textbf{0.089} \\
        Retrieval-only (CLIP) & 0.217 & 0.448 & 4.108 & 0.303 \\
        Retrieval (ImageNet) & 0.111 & 0.972 & 4.295 & 0.813 \\
        \bottomrule
        \end{tabular}
        }

    }
    \hfill
    \parbox[t]{.48\linewidth}{
        \centering
        
        \caption{\centering Ablation study on initialization settings.} 
        \label{tab:abla-init}
        
        \resizebox{7cm}{!}{
        \begin{tabular}{rccccc}
        \toprule
        Method  & mIOU $\uparrow$ & mCham $\downarrow$ & Skinning $\downarrow$ & Joint $\downarrow$\\
        \cmidrule(l){1-6}
        CASA (retrieval init) & \textbf{0.512} & 0.053 & 3.288 & \textbf{0.089} \\
        k-means rigging init & 0.305 & 0.452 & 3.336 & 0.659 \\
        Fixing skinning weight & 0.433 & \textbf{0.052} & \textbf{3.235} & 0.318 \\
        \bottomrule
        \end{tabular}
        }

        }
\end{table}



\begin{wrapfigure}{r}{0.4\textwidth}
    \vspace{-3mm}
    \includegraphics[width=0.4\textwidth]{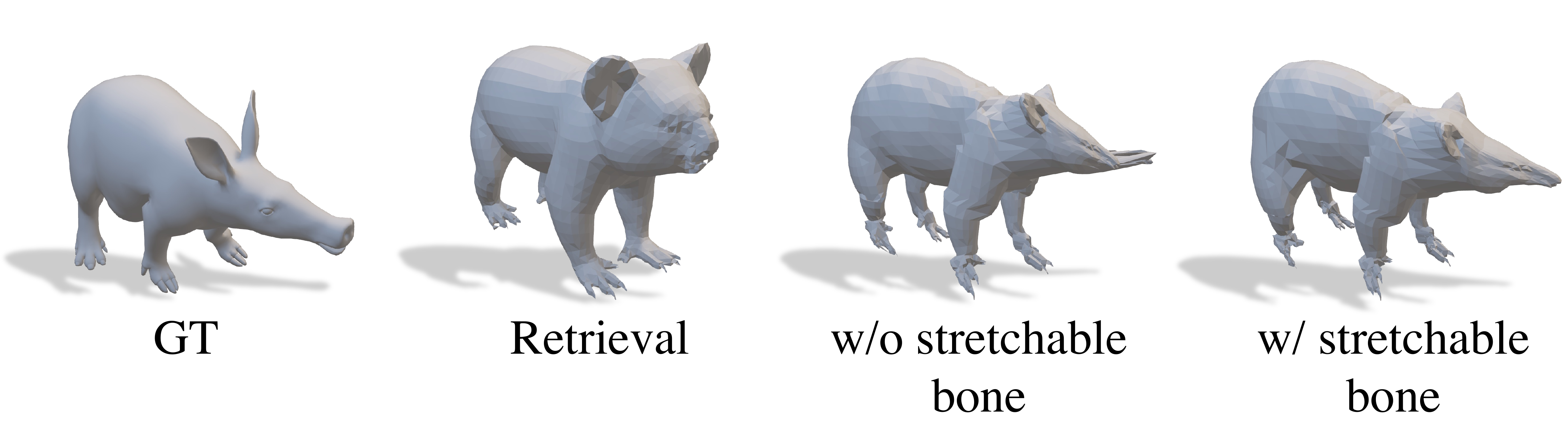}
  \vspace{-5mm}
  \caption{Stretchable bone}
  \label{fig:bone}
    \vspace{-5mm}
\end{wrapfigure}

    {\subsection{Ablation study}
We provide an ablation study to demonstrate the efficacy of each design choice in our framework. 

\textbf{Energy terms.} In Tab.~\ref{tab:abla-energy}, we ablate different terms of our energy function. Specifically, we consider the mask consistency, flow consistency, symmetry and smoothness regularization separately. 
We find that the mask part is crucial for the performance, while the optical flow part also improves the framework results. Removing the symmetry offset term results in performance degradation since this term plays a vital role in regularizing the neural offset, which can alleviate issues brought by the ambiguity of single-view monocular video. The smooth term leads to better qualitative results.

\textbf{Optimization.} We compare different optimization settings in Tab.~\ref{tab:abla-opti}. 
The results show that: 1) using neural offset field is superior to per-vertex offset or not deforming shape (CASA vs. w/o offset vs. w/o disp. field); stretchable skeleton helps (CASA vs. w/o skeleton); removing shape scaling step leads to performance degradation (CASA vs. w/o scaling). 

\textbf{Retrieval strategies.} In Tab.~\ref{tab:abla-retri}, we test different retrieval strategies. Our result demonstrates that CLIP is the preferred backbone for retrieval, most likely due to training with significantly richer semantic information than ImageNet pretrained models.

\textbf{Initialization.} We test different initialization strategies for skinning weights in Tab.~\ref{tab:abla-init}. We replace the rigging from the retrieved animal by using k-means for initializing weights. The comparison in the table shows that high-quality rigging weight initialization is essential for good shape predictions. The results of sphere initialization also confirm the necessity of the proposed retrieval phase.

\textbf{Stretchable bone.} 
Fig.~\ref{fig:bone} shows qualitative results with/without stretchable bone parameterization. Compared to merely deforming each vertex, stretchable bone deformation ensures global consistency and smoothness. As noted in the figure, without stretchable bones, we see discontinuity at the nose of the animal and a mismatch between the lower and upper mouth.
}

\subsection{Limitations}
While demonstrating superior performance in animal reconstruction, our method has a few remaining drawbacks. Firstly, the source asset bank restricts the diversity of the retrieved articulation topology. Bone length optimization partially alleviates such limitations through ``shrinking'' bones. Still, it cannot add new bones to the kinematic tree, e.g., we cannot create a spider from a quadrupedal template. Secondly, our method so far does not impose the constraint that bones are inside the mesh. We plan to tackle such challenges in the future.

\section{Conclusion}

In this paper, we propose CASA, a novel category-agnostic animation reconstruction algorithm. Our method can take a monocular RGB video and predict a 4D skeletal mesh, including the surface, skeleton structure, skinning weight, and the joint angle at each frame.  Our experimental results show that the proposed method achieves state-of-the-art performances in two challenging datasets. Importantly, we demonstrate that we could retarget our reconstructed 3D skeletal character and generate new animated sequences.

\textbf{Acknowledgements.}
The authors thank Vlas Zyrianov and Albert Zhai for their feedback on the writing. The project is partially funded by the Illinois Smart Transportation Initiative STII-21-07. We also thank Nvidia for the Academic Hardware Grant.

\clearpage
\bibliographystyle{plain}
\bibliography{main.bib}

\clearpage

\section*{Checklist}


\begin{enumerate}

\item For all authors...
\begin{enumerate}
  \item Do the main claims made in the abstract and introduction accurately reflect the paper's contributions and scope?
    \answerYes{}
  \item Did you describe the limitations of your work?
    \answerYes{}
  \item Did you discuss any potential negative societal impacts of your work?
    \answerYes{}
  \item Have you read the ethics review guidelines and ensured that your paper conforms to them?
    \answerYes{}
\end{enumerate}

\item If you are including theoretical results...
\begin{enumerate}
  \item Did you state the full set of assumptions of all theoretical results?
    \answerNA{}{}
        \item Did you include complete proofs of all theoretical results?
    \answerNA{}
\end{enumerate}

\item If you ran experiments...
\begin{enumerate}
  \item Did you include the code, data, and instructions needed to reproduce the main experimental results (either in the supplemental material or as a URL)?
    \answerNo{we will release the code and data upon acceptance}
  \item Did you specify all the training details (e.g., data splits, hyperparameters, how they were chosen)?
    \answerYes{}
        \item Did you report error bars (e.g., with respect to the random seed after running experiments multiple times)?
    \answerNo{}
        \item Did you include the total amount of compute and the type of resources used (e.g., type of GPUs, internal cluster, or cloud provider)?
    \answerYes{We use eight GTX 2080 Ti and RTX A6000 GPUs. }
\end{enumerate}

\item If you are using existing assets (e.g., code, data, models) or curating/releasing new assets...
\begin{enumerate}
  \item If your work uses existing assets, did you cite the creators?
    \answerYes{}
  \item Did you mention the license of the assets?
    \answerNo{}
  \item Did you include any new assets either in the supplemental material or as a URL?
    \answerNo{upon acceptance}
  \item Did you discuss whether and how consent was obtained from people whose data you're using/curating?
    \answerYes{BADJA is open-sourced; PlanetZoo has commercial license, following prior practices in our community~\cite{richter2017playing, caoHMP2020, hu2021sail3d}, we will ask users to purchase the game before accessing the dataset.}
  \item Did you discuss whether the data you are using/curating contains personally identifiable information or offensive content?
    \answerYes{Both datasets do not contain personal identifiable information. }
\end{enumerate}

\item If you used crowdsourcing or conducted research with human subjects...
\begin{enumerate}
  \item Did you include the full text of instructions given to participants and screenshots, if applicable?
    \answerNA{}
  \item Did you describe any potential participant risks, with links to Institutional Review Board (IRB) approvals, if applicable?
    \answerNA{}
  \item Did you include the estimated hourly wage paid to participants and the total amount spent on participant compensation?
    \answerNA{}
\end{enumerate}

\end{enumerate}


\clearpage
\appendix
    

\section{Additional Results}
\begin{figure*}[h]
    \centering  
    \includegraphics[width=1\linewidth]{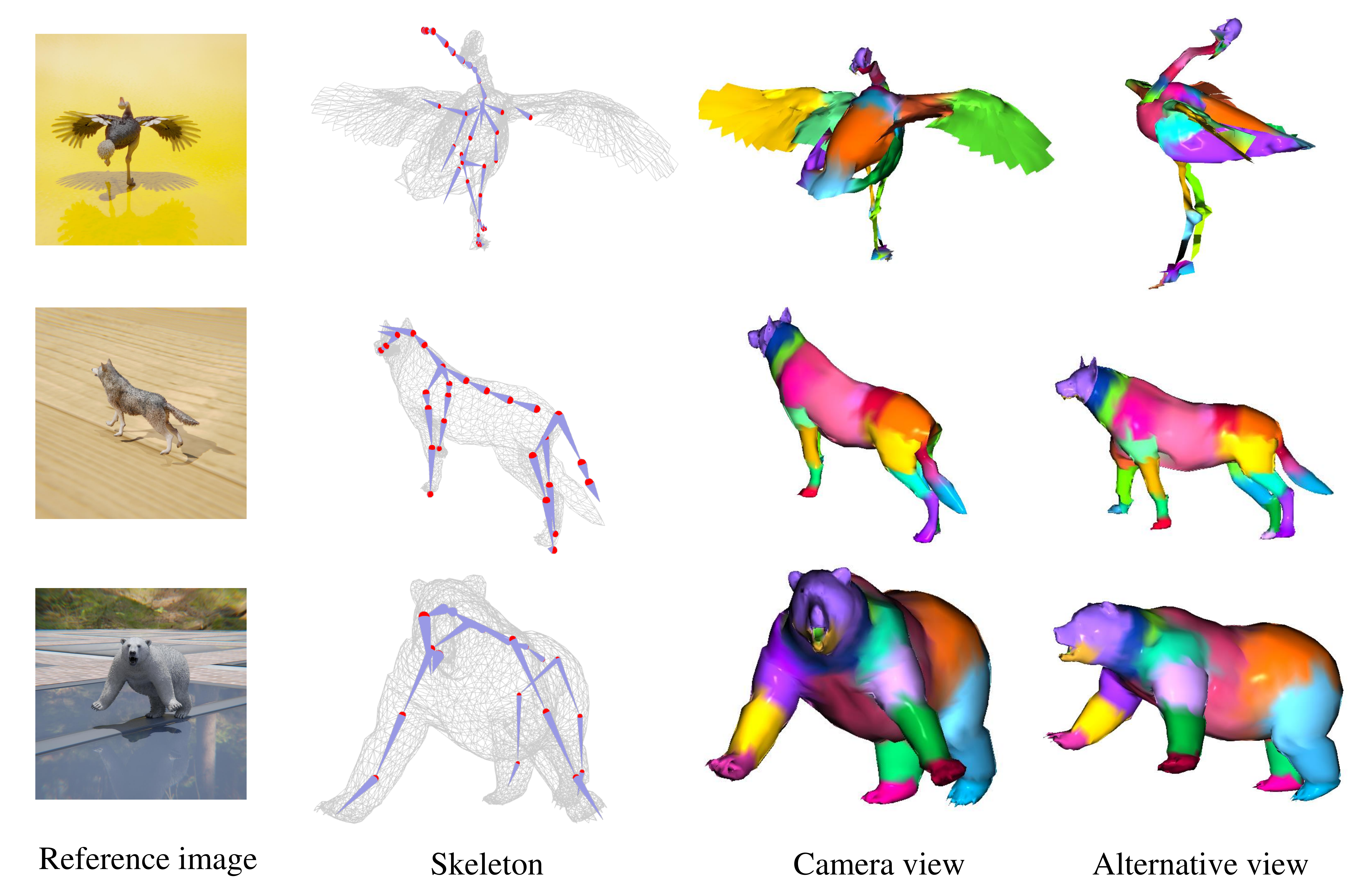}
    \caption{Additional qualitative results on \dataset{}.}
    \label{fig:additional}
\end{figure*}

We provide additional qualitative results on animals from \dataset{} in Fig.~\ref{fig:additional}. As shown in the figure, CASA well recovers the shape and topology of target animals from input videos. The sample of ostrich in this figure illustrates the limitations of CASA: 1) partial observation from monocular video leads to ambiguity on animal poses. The predicted shape fits well with the target from the camera view. However, from an alternative view, we can find that the wings of the prediction are of a wrong pose; 2) we do not impose a constraint on bones that they should be inside the mesh, resulting in joints in the head becoming outside the shape.

\begin{figure*}[h]
    \centering
    \includegraphics[width=1\linewidth]{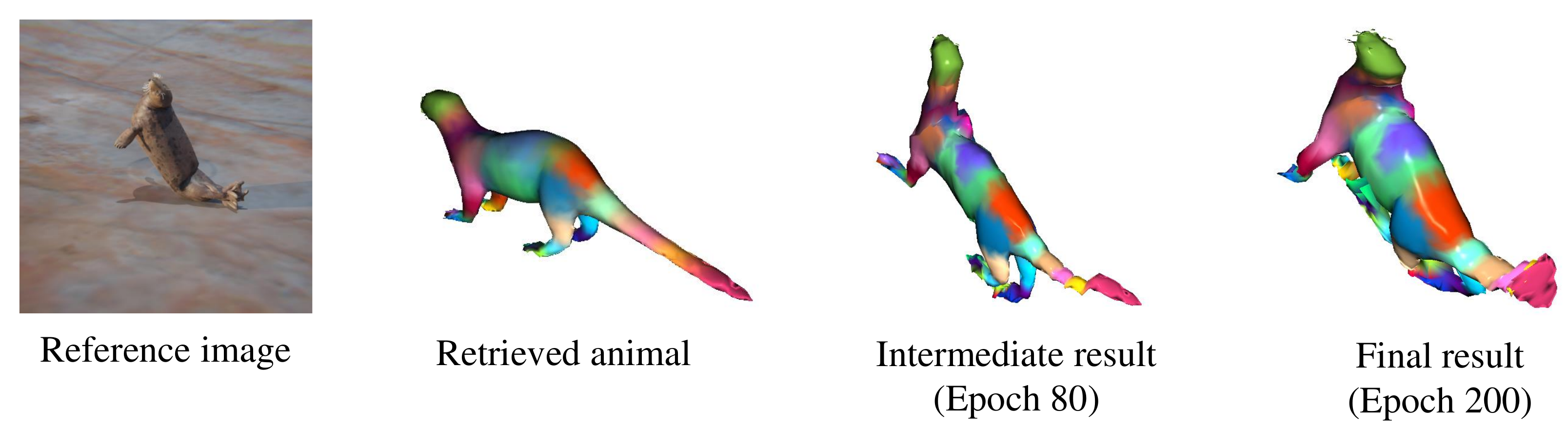}
    \caption{Given a retrieved animal with wrong articulation topology, CASA can recover a reasonable skeletal shape through our thanks to our stretchable bone formulation.}
    \label{fig:seal}
\end{figure*}

In order to demonstrate that CASA is able to adjust the topology of the initial shape, we provide the evolution of predictions for a seal in Fig.~\ref{fig:seal}. Although the retrieved animal is a quadrupedal animal (otter), CASA gradually warps the predicted shapes to the target seal, whose topology is significantly different from quadrupeds.

\section{Implementation Details}

We minimize the energy function using Adam optimizer~\cite{2014ADAM} with $\beta_1=0.9$ and $\beta_2=0.999$. We optimize the framework for $200$ epochs in total. A scheduling strategy is adopted. We optimize a scaling factor for the first 60 epochs by minimizing the mask term. This factor is initialized by aligning the bounding boxes between the rendering mask of the initial shape and the ground-truth mask. For the rest 140 epochs, we jointly optimize the bone length scale, joint angles, and the neural offset field. The learning rate is set to $5e-2$ for the first stage and $4e-3$ for all parameters in the second stage, except for the mesh scale and neural offset, which are set to $1e-3$. The trade-off weights for the mask term, flow term, smoothing term, and symmetry term are empirically set to $1e4$, $1e6$, $1e6$, and $1e4$, respectively. The neural offset field is parameterized by a four-layer MLP.

\section{Retrieval Result}
\begin{figure*}
\centering
\subfigure{
\begin{minipage}[b]{0.235\linewidth}
\includegraphics[width=1.\linewidth]{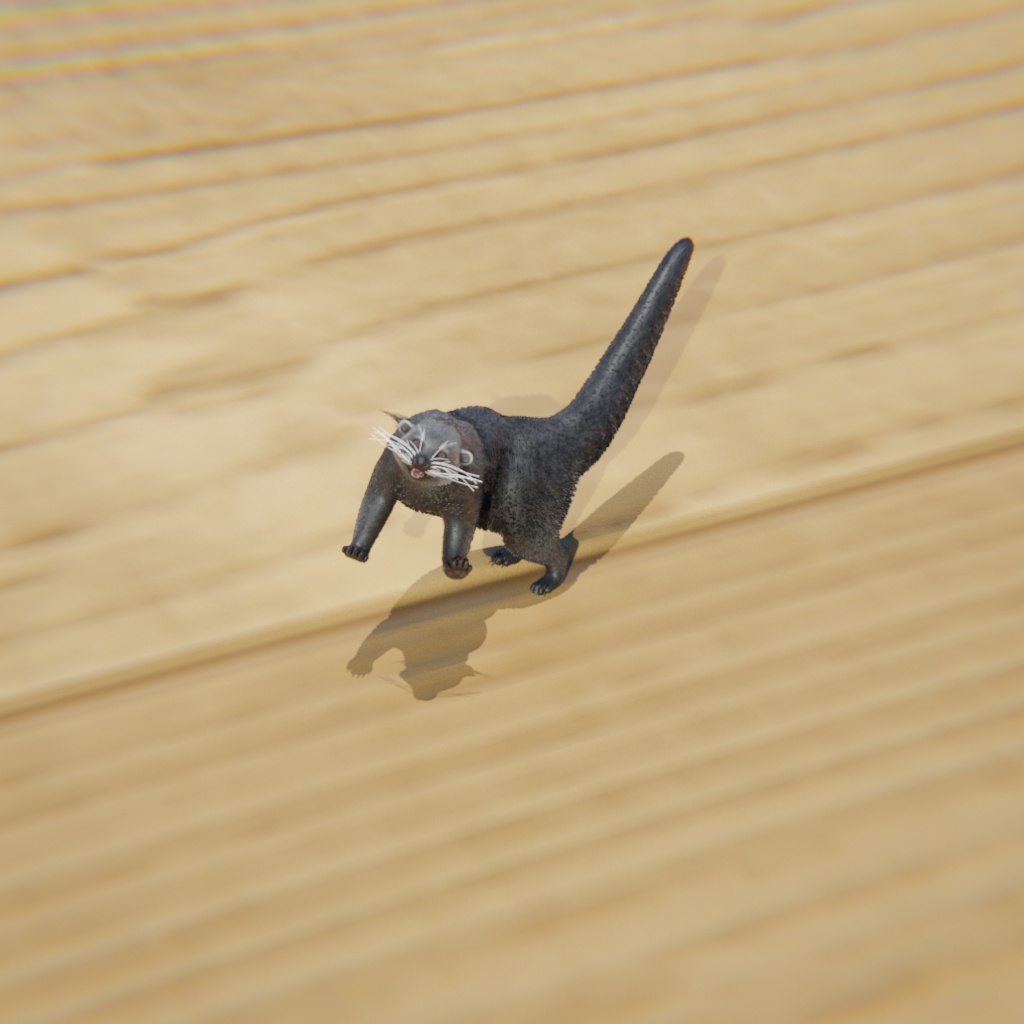}
\includegraphics[width=1.\linewidth]{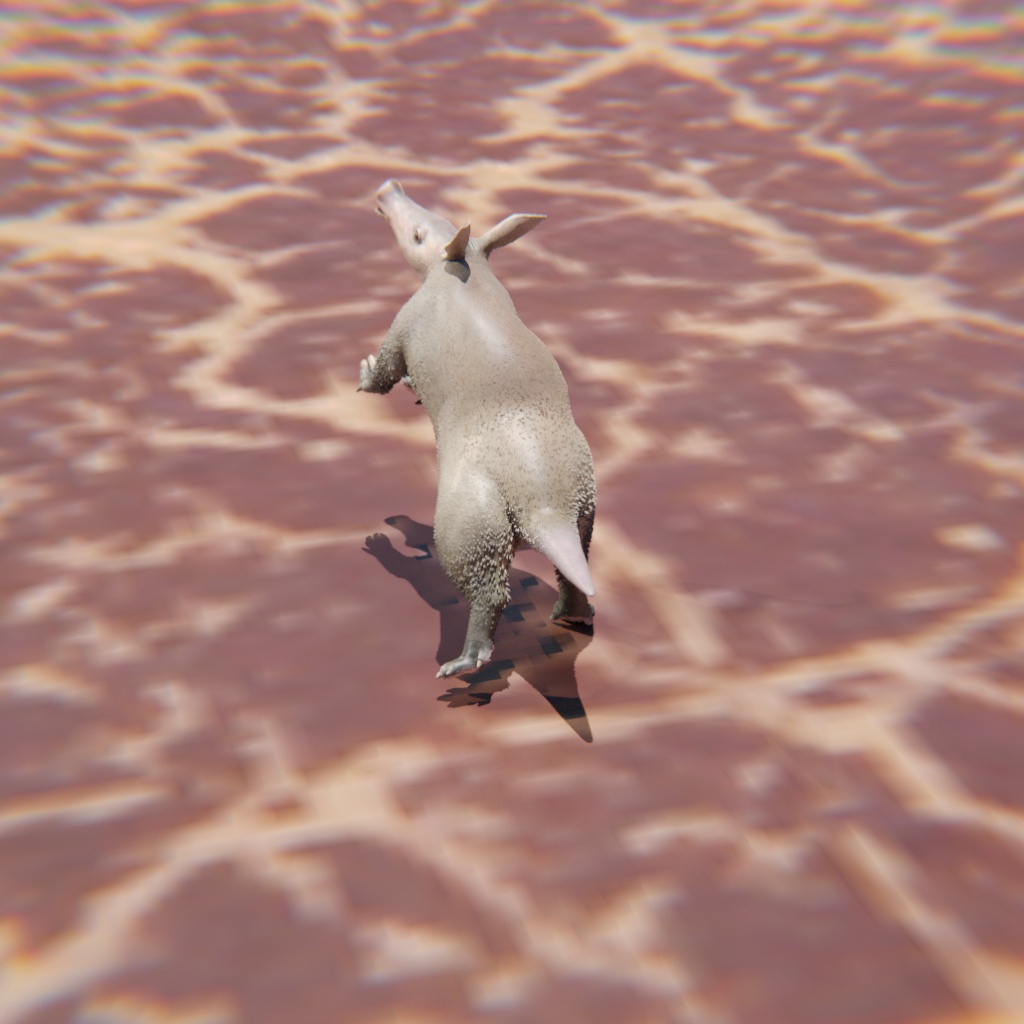}
\includegraphics[width=1.\linewidth]{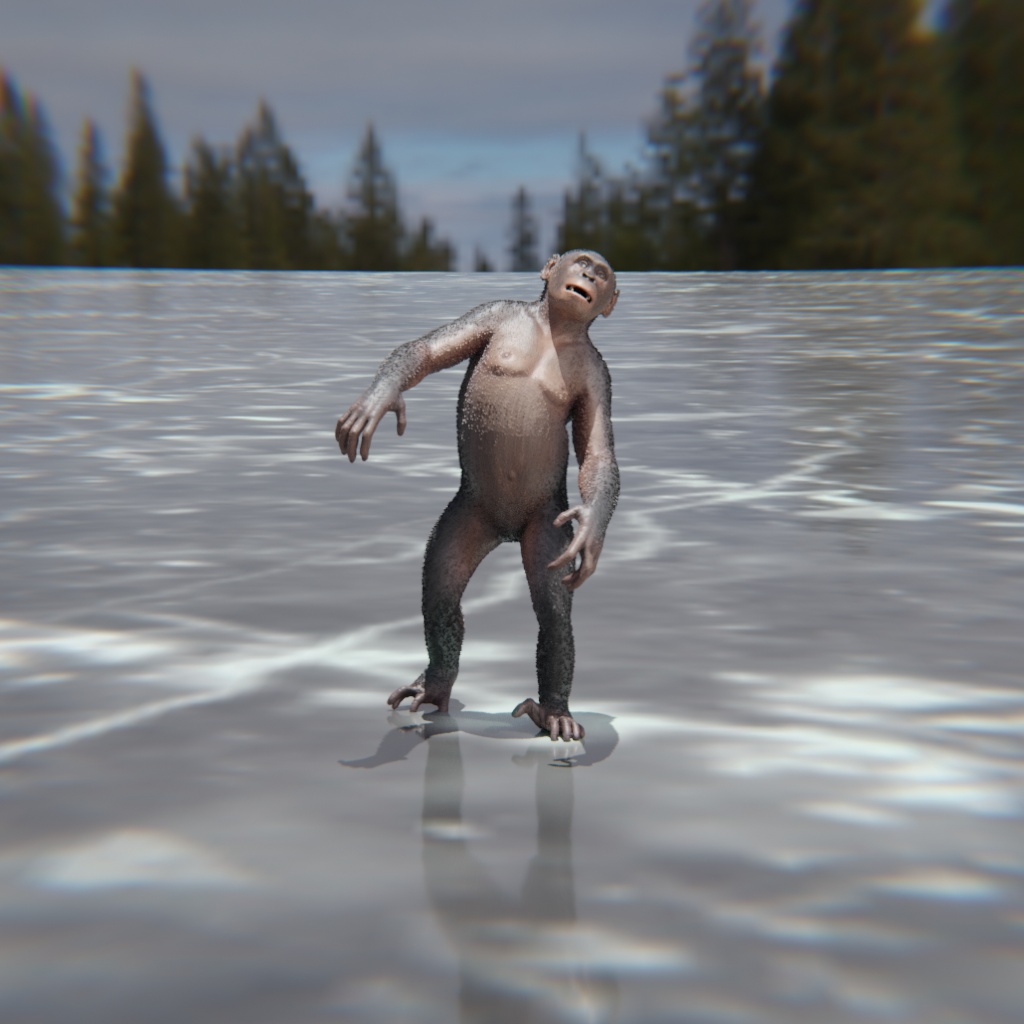}
\includegraphics[width=1.\linewidth]{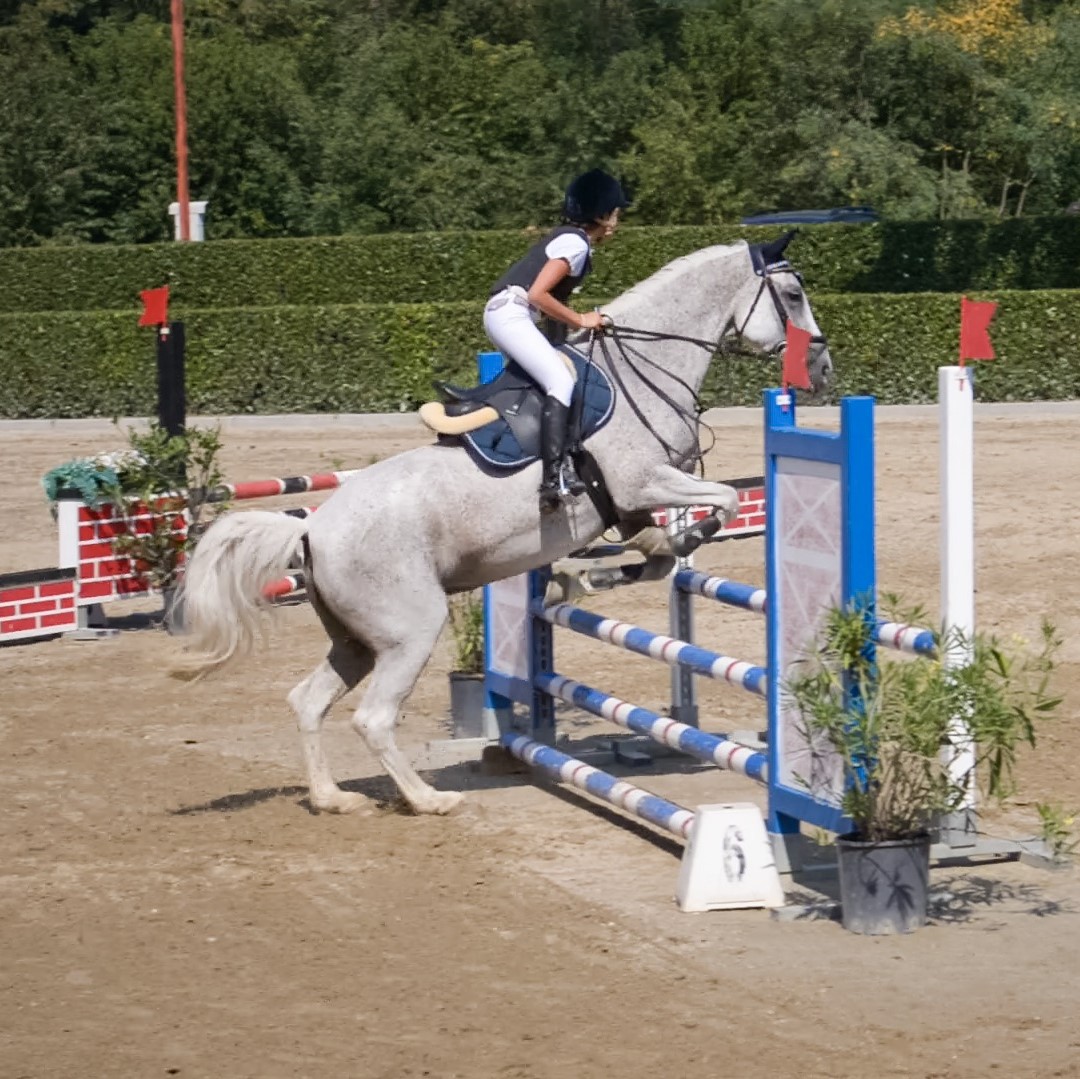}
\caption*{Input animal}
\end{minipage}}
\subfigure{
\begin{minipage}[b]{0.235\linewidth}
\includegraphics[width=1.\linewidth]{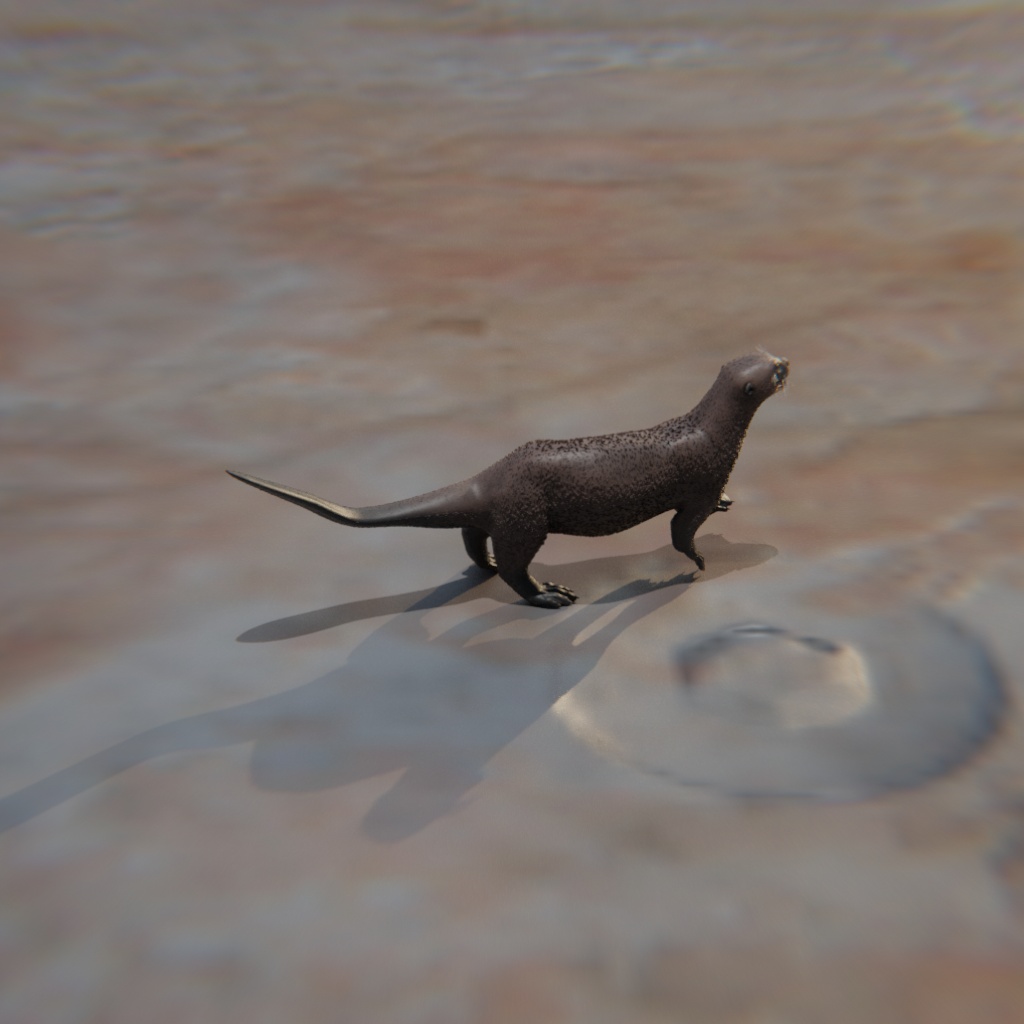}
\includegraphics[width=1.\linewidth]{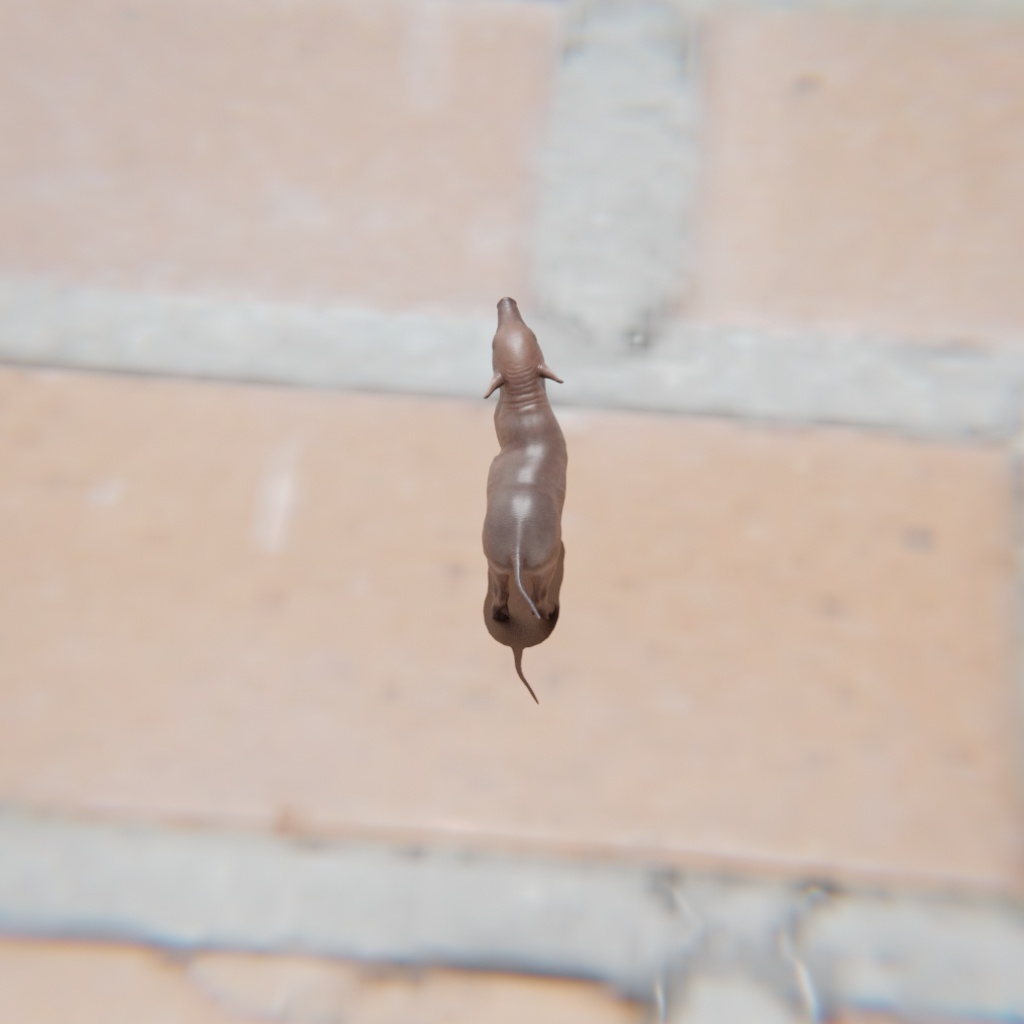}
\includegraphics[width=1.\linewidth]{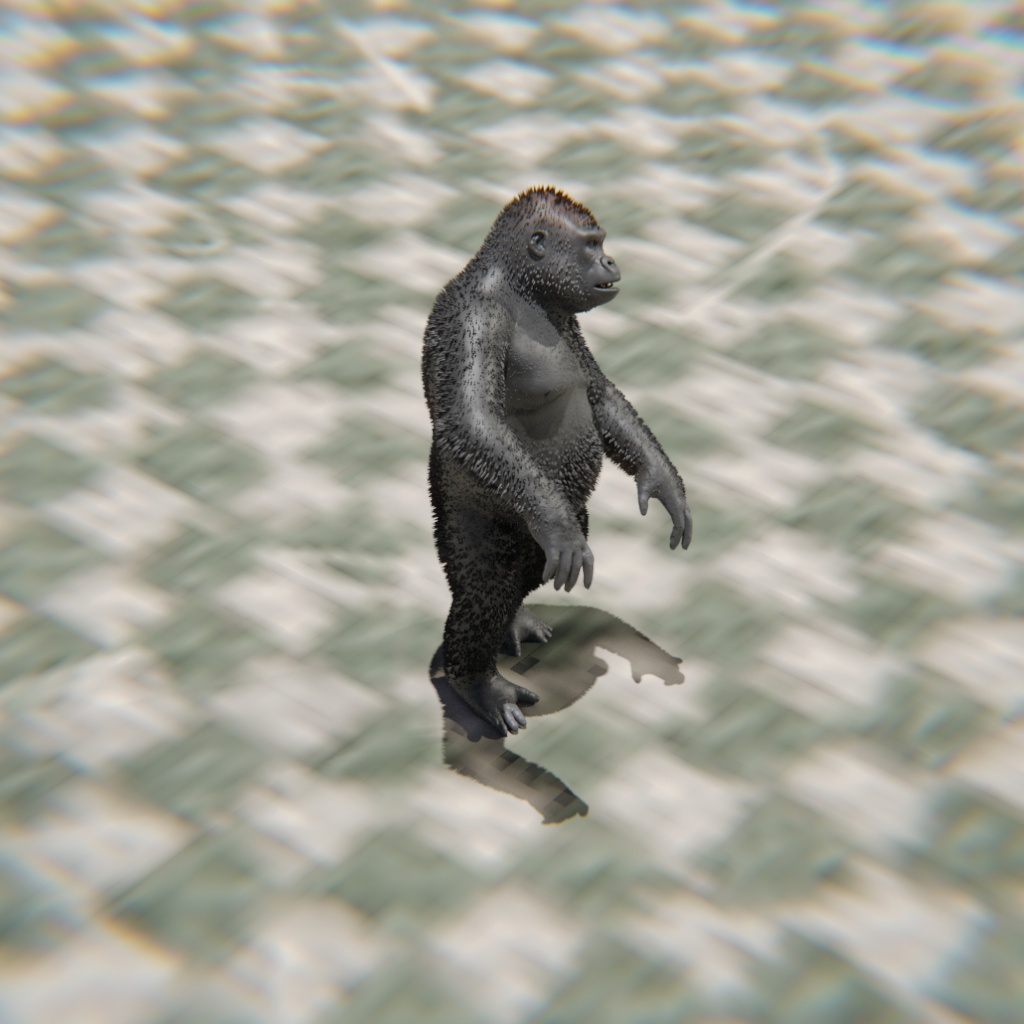}
\includegraphics[width=1.\linewidth]{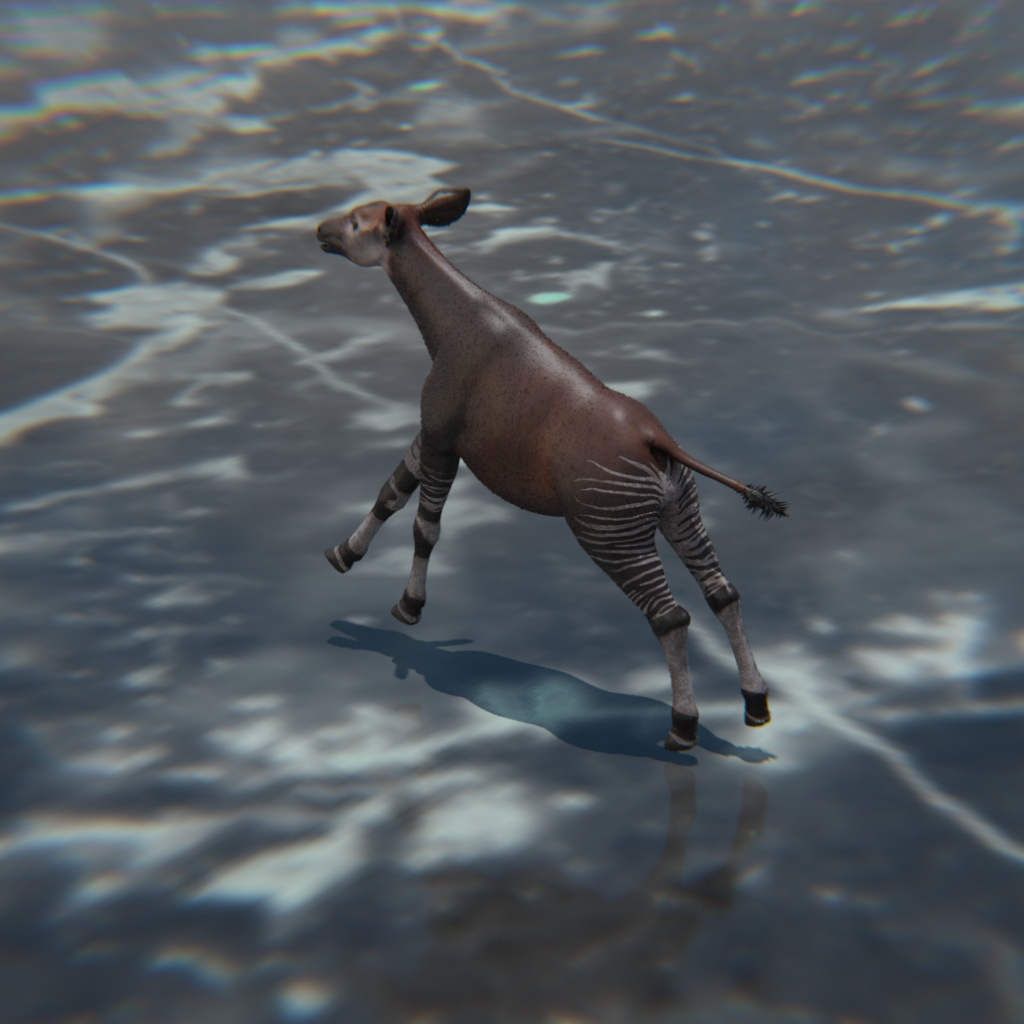}
\caption*{Retrieved animal}
\end{minipage}}
\subfigure{
\begin{minipage}[b]{0.235\linewidth}
\includegraphics[width=1.\linewidth]{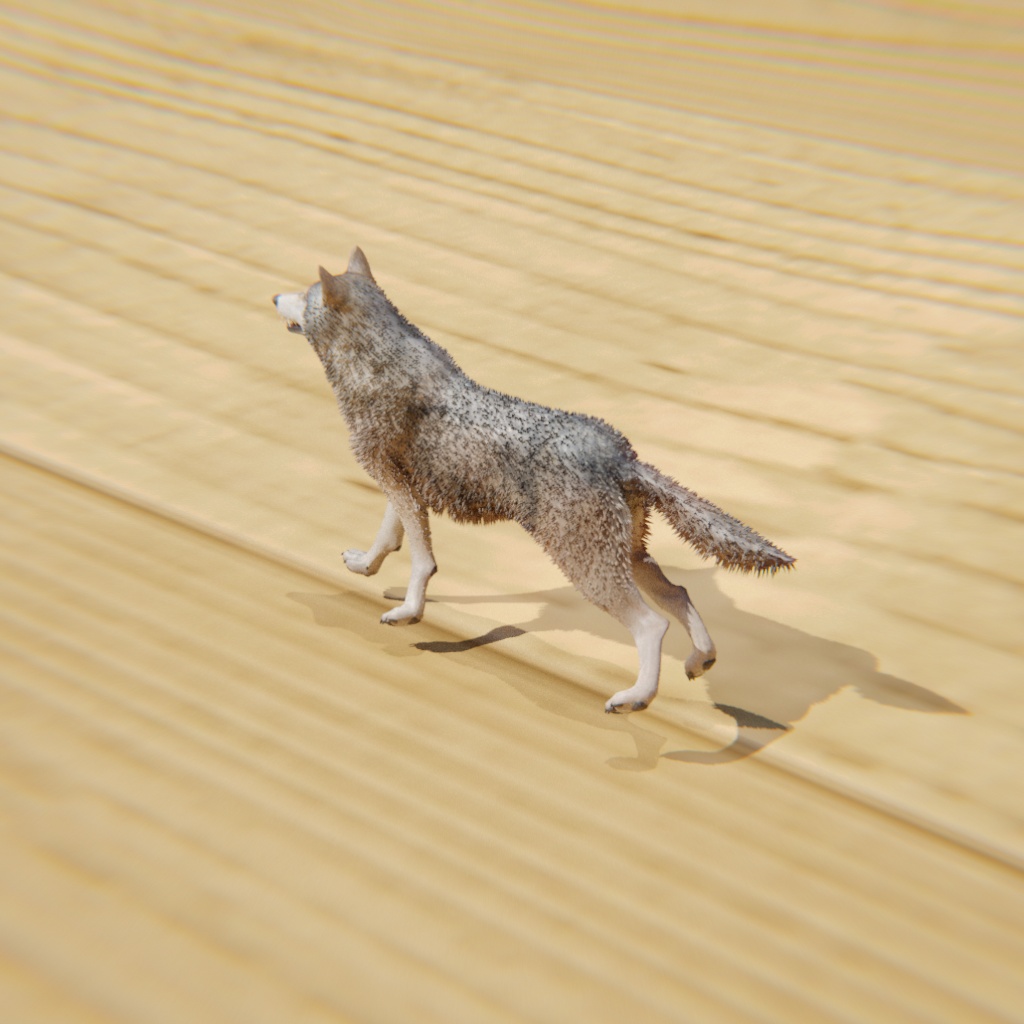}
\includegraphics[width=1.\linewidth]{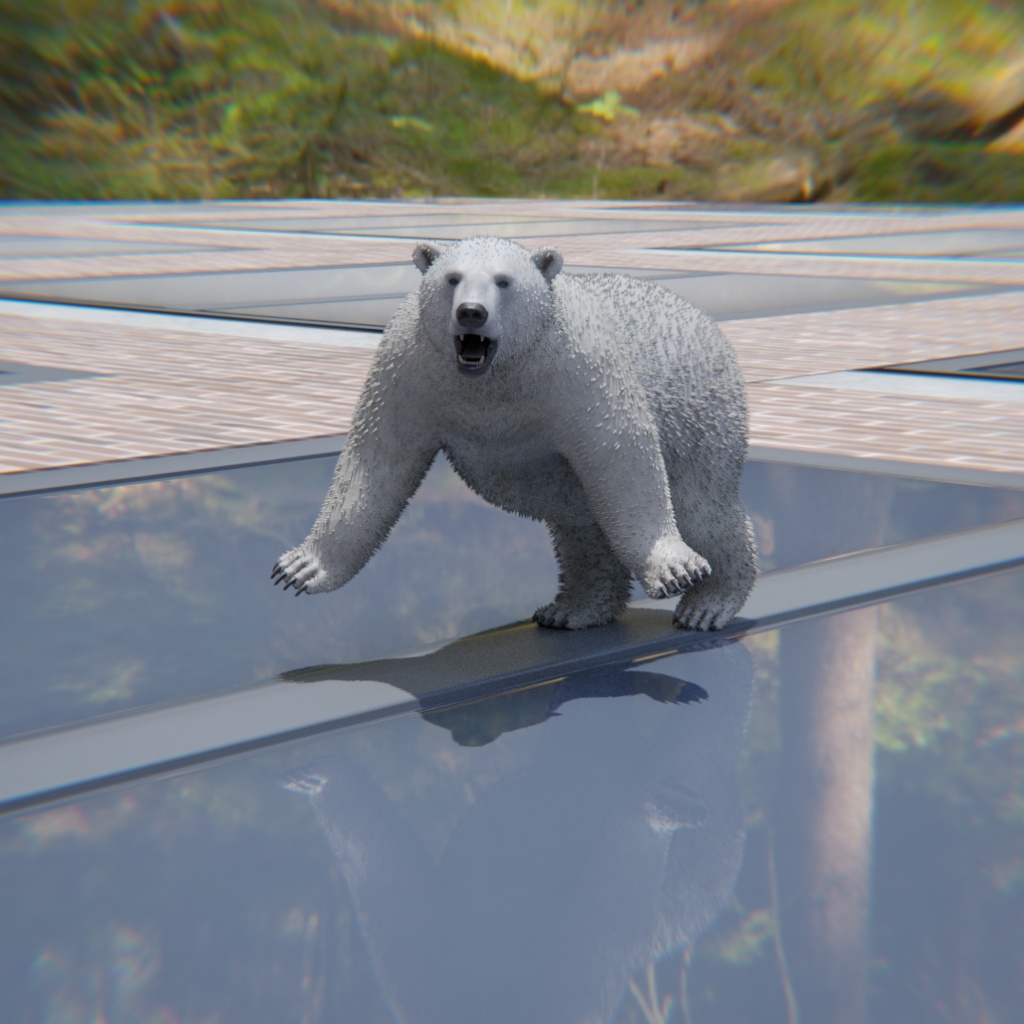}
\includegraphics[width=1.\linewidth]{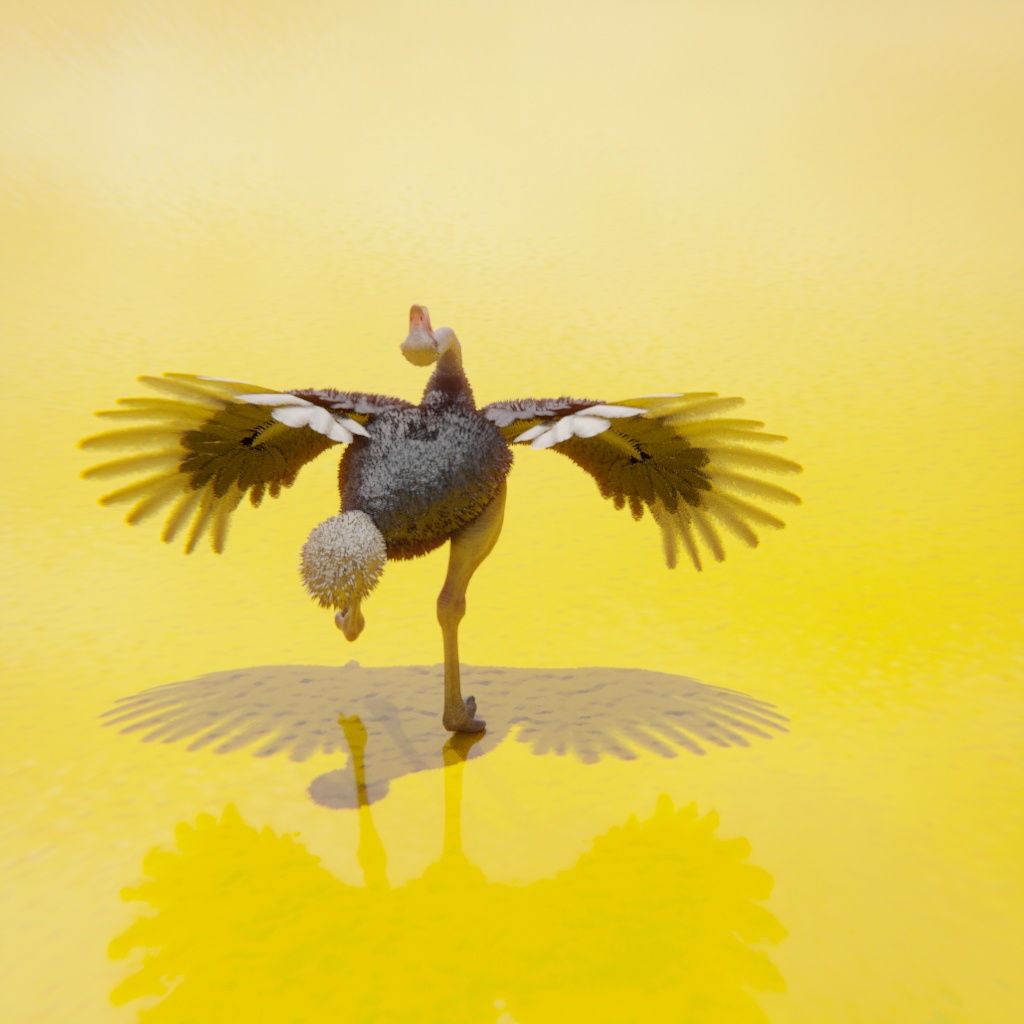}
\includegraphics[width=1.\linewidth]{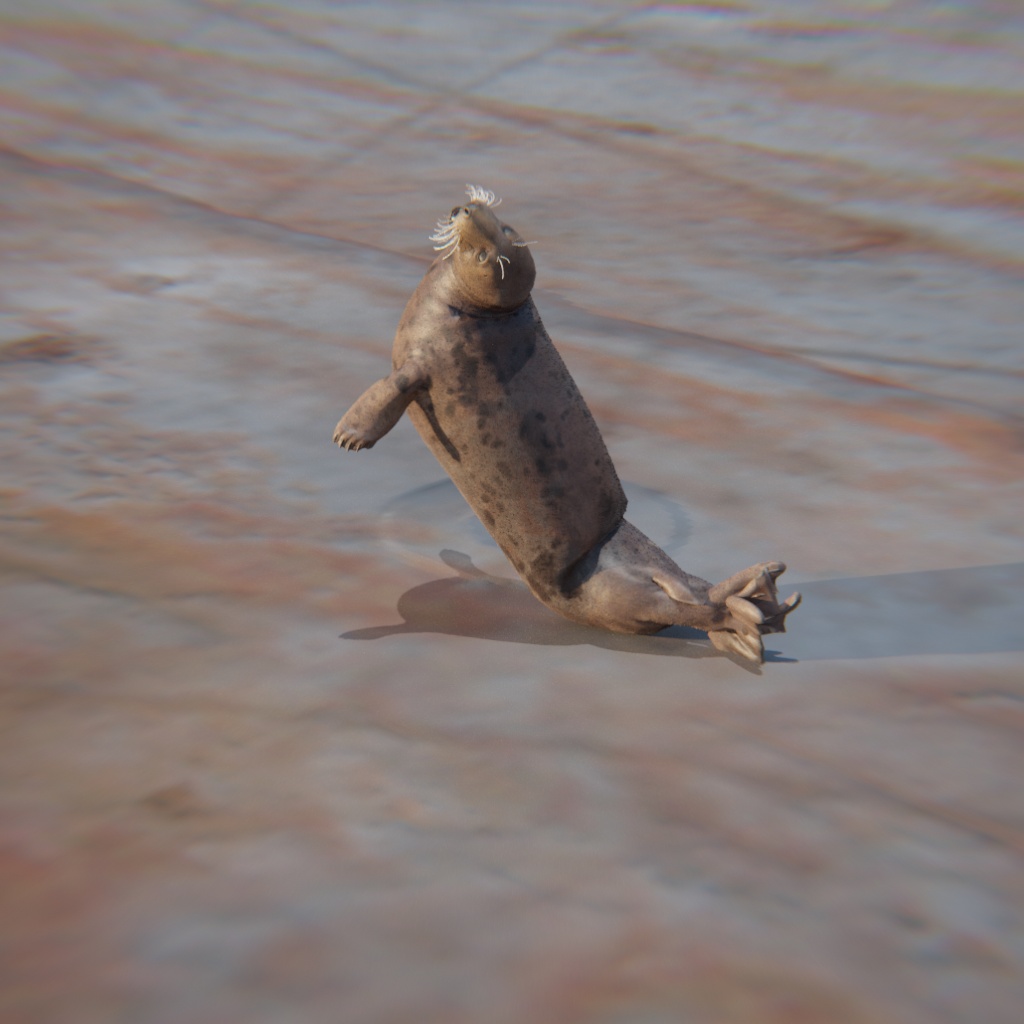}
\caption*{Input animal}
\end{minipage}}
\subfigure{
\begin{minipage}[b]{0.235\linewidth}
\includegraphics[width=1.\linewidth]{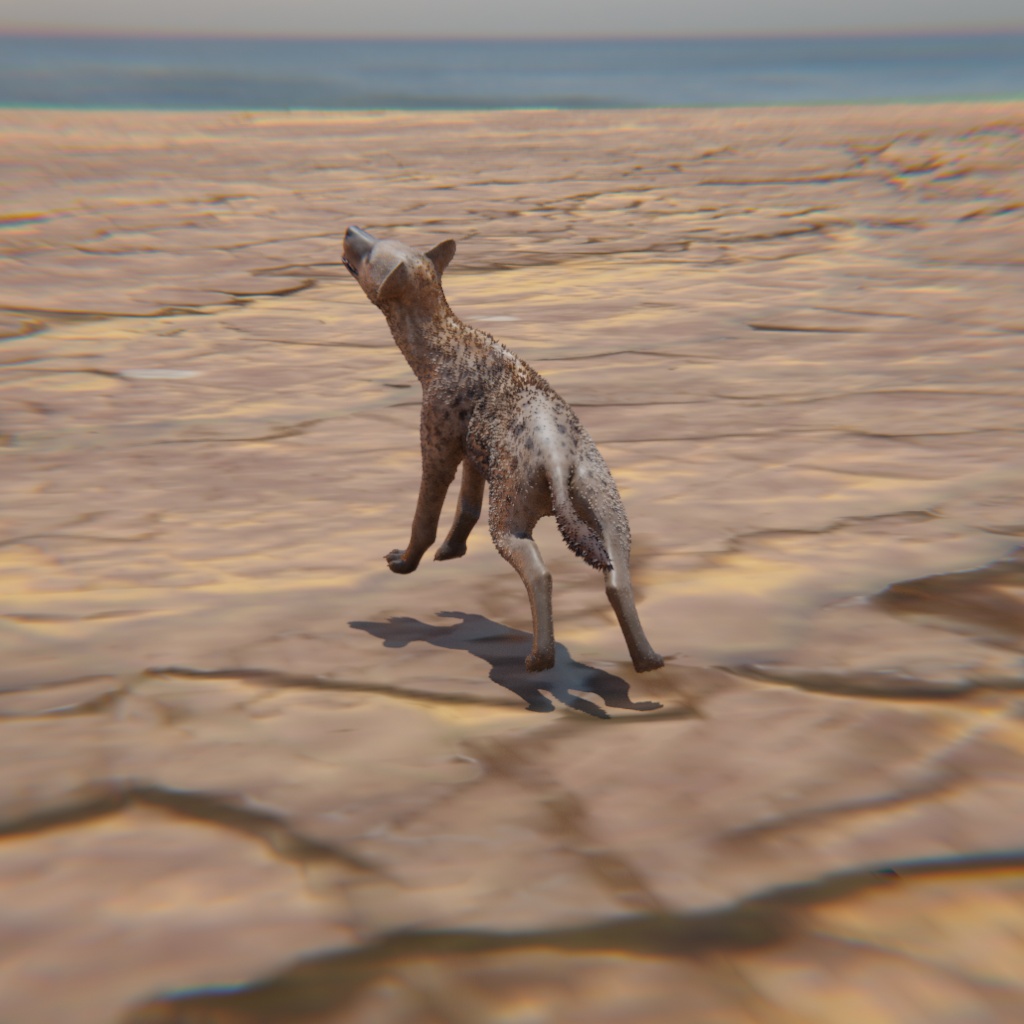}
\includegraphics[width=1.\linewidth]{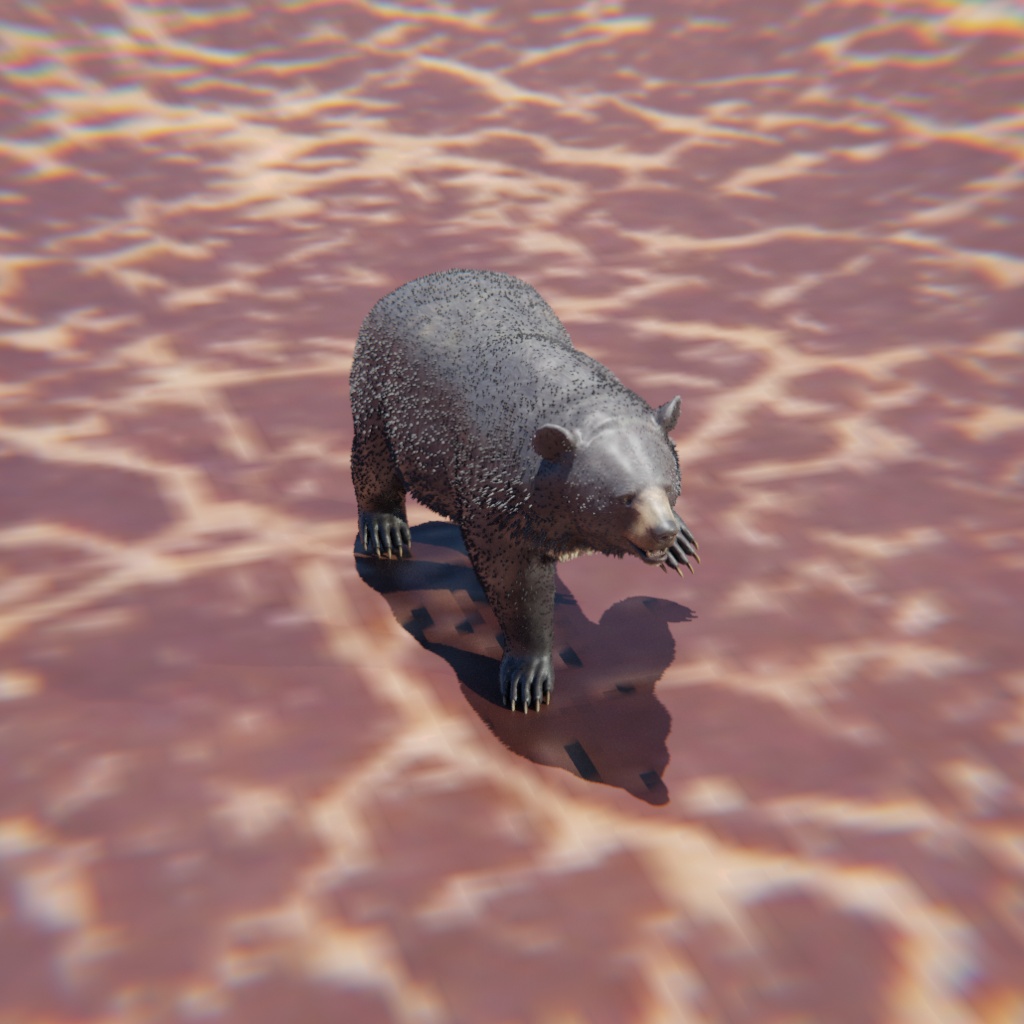}
\includegraphics[width=1.\linewidth]{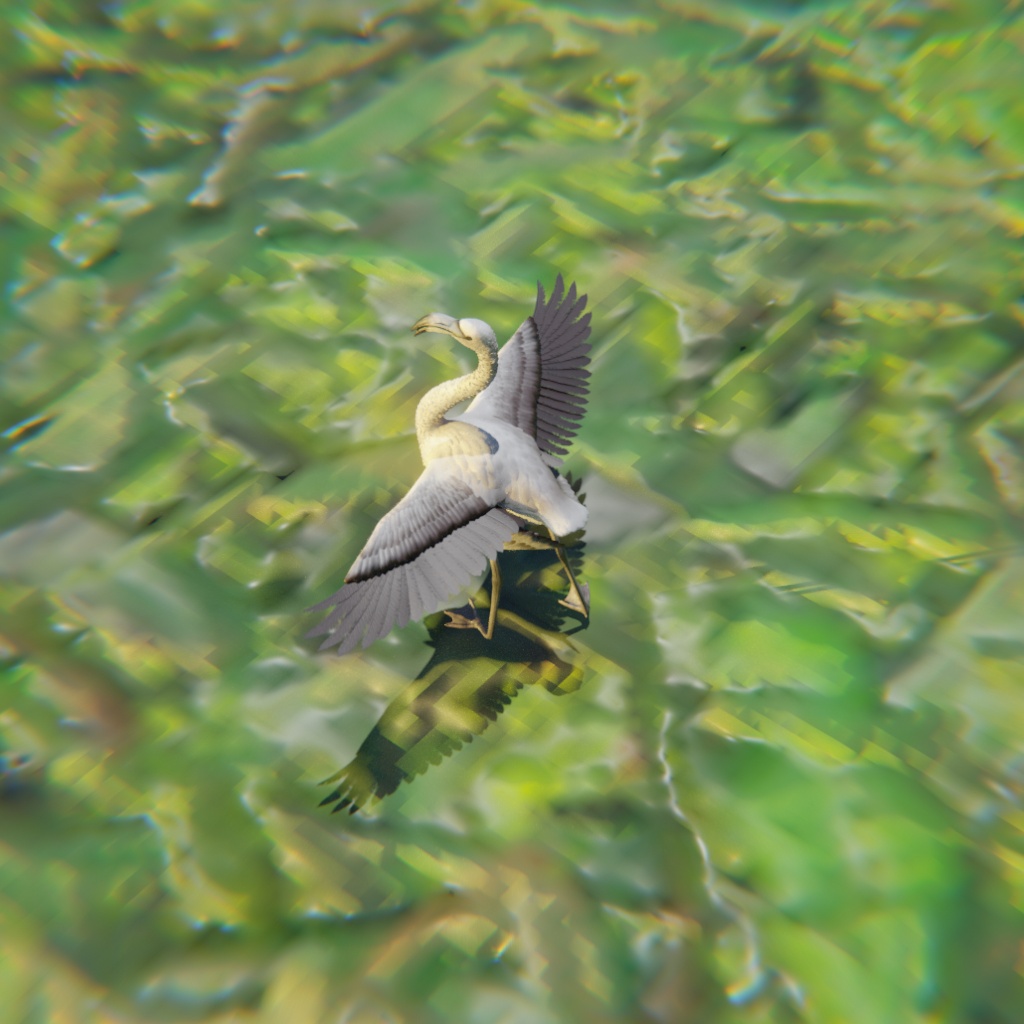}
\includegraphics[width=1.\linewidth]{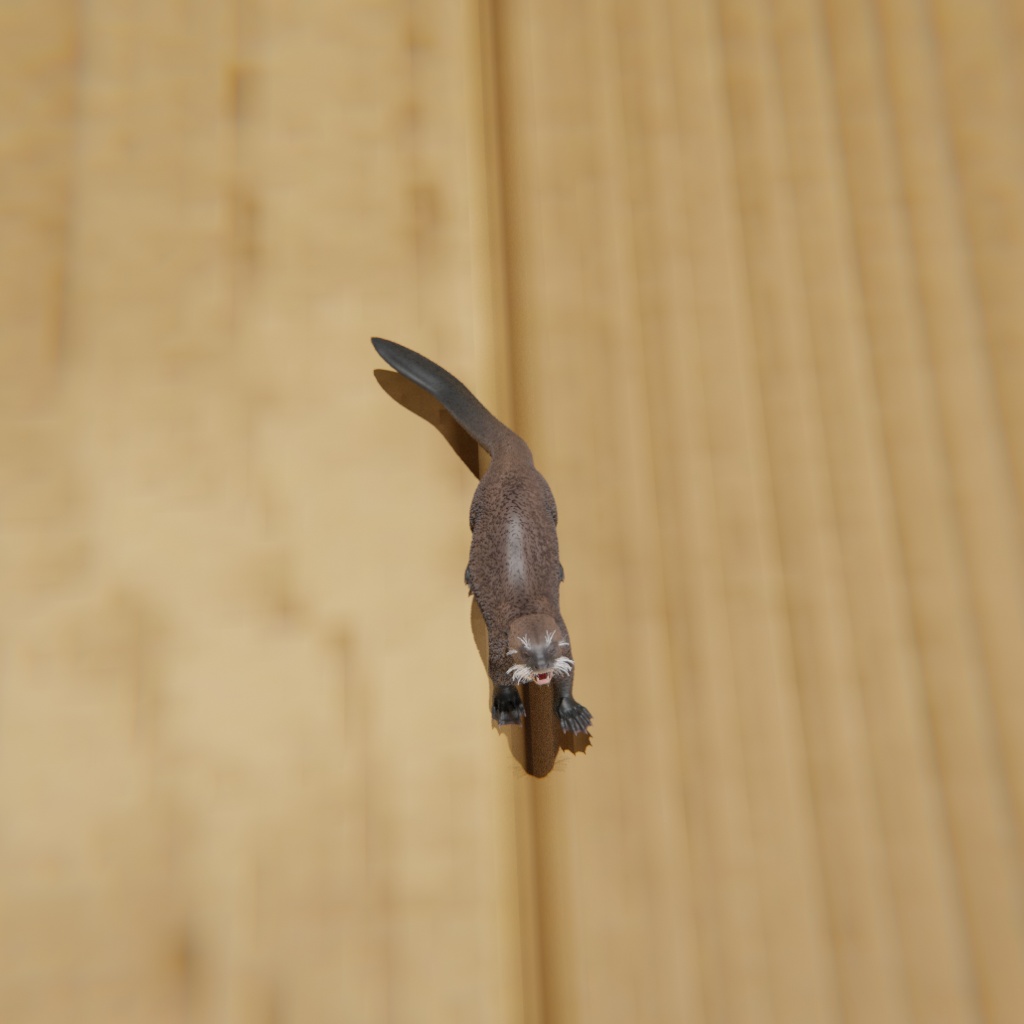}
\caption*{Retrieved animal}
\end{minipage}}
\caption{Retrieval results.}
\label{fig:retrieve}
\end{figure*}

Fig.~\ref{fig:retrieve} shows some retrieval results. Input animals are of various colors, skinning textures, and poses, together with different backgrounds and lighting conditions in their corresponding videos. Still, our method can retrieve animals from the database with a similar topology to the input, as CLIP~\cite{wang2021clip} captures high-level semantic information and avoids distractions brought by different appearances and background environments. The sample of the seal at the bottom right is a failure case, as the two animals have different typologies. However, as shown in Fig.~\ref{fig:seal}, CASA is able to bridge the gap between retrieved and target animals.

\section{Stretchable Bone Formulation} To incorporate stretchable bone parameterization into our linear blend skinning model, we have to re-formulate the skinning equation. The original formulation for linear blend skinning is defined as:
\begin{equation}
    \mathbf{v}_i^t= \sum_{k = 0}^K w_{i,k} \mathbf{T}_k^t \mathbf{v}_i
\end{equation}
where $\mathbf{v}_i^t$ and $\mathbf{v}_i$ are the position of vertex $i$ at frame $t$ and the rest pose, respectively. $w_{i,k}$ is the rigging weight. $\mathbf{T}_k^t$ is the transformation of bone $k$ at frame $t$.

Since the bone transformation $\mathbf{T}_k^t$ is rigid, we can replace it by a translation and a rotation. In particular, given the head positions of each bone before and after deformation (which can be calculated by performing transformations according to forward kinematics), we can define the skinning model as:
\begin{equation}
    \mathbf{v}_i^t= \sum_{k = 0}^K w_{i,k} (\mathbf{c}_i^t+ \mathbf{R}_k^t(-\mathbf{c}_i+\mathbf{v}_i))
\label{eq:LBS2}
\end{equation}
where $\mathbf{c}_k$ and $\mathbf{c}_k^t$ are the head positions of bone $k$ before and after deformation. $\mathbf{R}_k^t$ is the rotation matrix decomposed from the rigid transformation $\mathbf{T}_k^t$. 

Simply adding a scaling term into the skinning model can result in shape explosion, especially for vertices located near tails of bones, as illustrated in~\cite{2011stretch}. Following the solution proposed in~\cite{2011stretch}, we define \textit{endpoint weight functions} $e_k(\mathbf{p})$ for all bones, to control shape deformation in scaling. With the functions, the new skinning model is defined as:
\begin{equation}
    \mathbf{v}_i^t= \sum_{k = 0}^K \mathbf{w}_{i,k} (\mathbf{c}_k^t + R_k^t (e_k(\mathbf{p}_i)\mathbf{s}_k + (\mathbf{p}_i - \mathbf{c}_k))
\label{lbs-stretch}
\end{equation}
where $\mathbf{s}_k = (\mathbf{b}_k - 1)(\mathbf{d}_i - \mathbf{c}_i)$ is the full stretching vector at the tail of bone $\mathbf{d}_i$, $\mathbf{b}_k$ is the bone scaling factor. Each vertex $\mathbf{p}_i$ should have a unique stretching vector corresponding to this bone, which is decided by the endpoint weight function $e_k(\mathbf{p}_i)$.

Typically the value of endpoint weight functions increases from 0 to 1 along the rest vector $(\mathbf{d}_k - \mathbf{c}_k)$ of each bone. We define these functions by projecting each vertex $\mathbf{v}_i$ onto the nearest point of each bone, calculating the fraction of where it falls between the head and tail of the bone as the weight:
\begin{equation}
    e_k(\mathbf{v}_i) = \frac{\|\textrm{proj}_k(\mathbf{v}_i) - \mathbf{c}_k\|}{\|\mathbf{d}_k-\mathbf{c}_k\|}
\end{equation}
where $\mathbf{c}_k$ and $\mathbf{d}_k$ are the head and tail positions of bone $k$, $\textrm{proj}_k(\mathbf{v}_i)$ is the projection of $\mathbf{v}_i$ onto the bone.

\section{Template-based Baseline}
We compare CASA with an additional template-based baseline in Tab.~\ref{tab:acfm}. In order to ensure a fair comparison, we only incorporate quadruped animals in the PlanetZoo testset in experiments since ACFM is a category-specific method. The results show that CASA outperforms ACFM without pre-training on a large-scale database.

\begin{table}[h]

    \centering
    \bigskip
    \centering
    
    \vspace{-5mm}
    \caption{Quantitative results on quadruped animals in PlanetZoo testset. }
    \label{tab:acfm}
    \begin{tabular}{rcc}
    \toprule
    Method  & mIOU $\uparrow$ & mCham $\downarrow$ \\
    \cmidrule(l){1-3}
    A-CFM~\cite{kokkinos2021learning} & 0.234 & 0.246 \\
    CASA & 0.499 & 0.050 \\
    \bottomrule
    \end{tabular}
\end{table}

\section{Optimization Framework Comparison}
We compare CASA-retrieval + LASR with the entire CASA pipeline. For LASR, rather than using the whole coarse-to-fine optimization steps, we utilized only the final step. The quantitative comparison can be found in table \ref{tab:optimize}.

\begin{table}[h]
    \centering
    \bigskip
    \centering
    
    \vspace{-5mm}
    \caption{Optimization Comparison on PlanetZoo testset. }
    \label{tab:optimize}
    \begin{tabular}{rcccc}
    \toprule
    Method  & mIOU $\uparrow$ & mCham $\downarrow$ & Skinning $\downarrow$ & Joint$\downarrow$\\
    \cmidrule(l){1-5}
    Retrieval+LASR & 0.191 & 0.158 & \textbf{2.975} & 0.407 \\
    CASA & \textbf{0.512} & \textbf{0.053} & 3.288 & \textbf{0.089} \\
    \bottomrule
    \end{tabular}
\end{table}

\section{Efficiency}
We conduct a comparison on the efficiency between  CASA and the other two state-of-the-art methods, LASR~\cite{2021LASR} and ViSER~\cite{2021VISER}. We test their average running time for animals in the testing set of PlanetZoo, using one RTX 2080Ti. The results are shown in Tab.~\ref{tab:time}. Our method is approximately 3$\times$ faster than LASR and 5$\times$ faster than ViSER, demonstrating that our method is highly efficient. This is natural as CASA do not need to train any complex neural networks.

\begin{table}[h]
    \centering
    \bigskip
    \centering
    \vspace{-5mm}
    \caption{Running time (min)} 
    \label{tab:time}
    \begin{tabular}{rccccc}
    \toprule
    Method  & ViSER & LASR & CASA \\
    \cmidrule(l){1-6}
    Time & 234.2 & 144.6 & 46.7 \\
    \bottomrule
    \end{tabular}
\end{table}

\section{Detailed Implementation of Retrieval}
We denote the input monocular video containing an animal to be reconstructed as $\{I_t\}_{t=1,...,T_1}$. The 3D asset bank for retrieving includes $N$ categories of animals. For each animal $i$, the asset bank includes a 3D mesh $M^i$, a skeleton $S^i$ and a realistic rendering video $\{I_t\}_{t=1,...,T_2}$. In our retrieval stage, our goal is to search the animal with the best similarity to the input animal in the asset bank, and the output should be its 3D shape $\mathbf{s}_j$. In short, our retrieval stage learns a mapping function from 2D video to 3D mesh. 
Since the input and output of retrieval are in different modalities, the target mapping function is not easy to learn. We propose to measure the similarity of the input video $\{I_t\}$ and all rendering videos in the asset bank using CLIP, obtain the most similar rendering video $\{I^r_t\}$, and take the corresponding animal mesh $M^r$ as the retrieval result.

To be specific, given the input video and any animal shape in the asset bank, we utilize pre-trained CLIP to encode all frames of the input video and the photo-realistic rendering :

$$
\{F^\mathrm{input}_t\}_{t=1,...,T_1} = \{g_\mathrm{CLIP}(\{I_t\}\}_{t=1,...,T_1} 
$$
$$
\{F^{j}_v\}_{v=1,...,V} = \{g_\mathrm{CLIP}(\pi(\mathbf{s}_j, \mathbf{q}_v)\}_{v=1,...,V}
$$

where $\{I_t\}\}_{t=1,...,T_1}$ is the input video, $\mathbf{s}_j$ is the $j$-th animal shape, $g_\mathrm{CLIP}$ is the image embedding network of the CLIP model and $\pi(\mathbf{s}_j, \mathbf{q}_v)$ is the photo-realistic rendering of the articulated shape $\mathbf{s}_j$ at a randomized skeletal poses $\mathbf{q}_v$. $T_1$ is the length of the input video. $V$ is the number of randomized skeletal pose. Note that we render only one frame for each pose.

With the corresponding encoding features $\{F^\mathrm{input}_t\}$ and $\{F^j_v\}$, we first calculate the similarity $D$ between the $t$-th input frame and the $v$-th skeletal pose for the $j$-th shape by:

$$
S(I_{t}, \pi(\mathbf{s}_j, \mathbf{q}_v)) = \|F^\mathrm{input}_{t} - F^j_{v}\|^2_2
$$

With the similarity defined above, We are able to retrieve the animal category $r$ in the asset bank that is the most similar with the input animal, which is formulated as:


$$
r = \arg\max_j  \max_t \max_v S(I_{t}, \pi(\mathbf{s}_j, \mathbf{q}_v))
$$

We then obtain the 3D animal shape $\mathbf{s}_r$ together with its corresponding skeleton and skinning weight from the asset bank, which is used for initializing the following optimization process.

\section{Initialization Strategy of CASA}
We provide details of our initialization strategy for CASA. 

In the skeletal optimization stage of CASA, optimizing parameters include a displacement field, the skinning weight, joint rotation angles for each bone, a root transformation, and bone length scales. During initialization, the displacement field is defined by a Multi-Layer Perception (MLP) with random network parameters. The skinning weight is set according to the retrieval results. Bone length scales are all set to 1. The joint angles are set to T-pose. 

The root transformation is represented as a combination of rotation and translation. For the convergence of skeletal optimization, the root transformation should ensure the consistency of the mesh in the camera coordinate with the ground-truth. Instead of estimating such transformation in world coordinate, we propose to predict camera parameters for each video, and fix the root transformation as an identical one. 
In practice, for synthetic videos in which the camera parameters are available, we directly use them as initialization. For videos where camera parameters are not available, we adopt a naive strategy to predict the camera parameters. 

Specifically, we first optimize the camera intrinsic and focus point offset from the origin by minimizing the mask loss between rendering and ground-truth. In each iteration, we randomly sample a point as the location of the camera from a sphere with its center at the origin and radius $R$ set according to the animal scale. 

When meeting the convergence criteria, we fix the camera intrinsic and focus point offset. We then pick camera locations sampled before with the Top-10 lowest mask loss values. For each location, we randomly sample points in its nearby regions as new camera locations, calculate the corresponding mask loss values, and select the location with the lowest value as our final camera location.

\section{More information on PlanetZoo}
\begin{figure}[h]
\centering
\includegraphics[width=\linewidth]{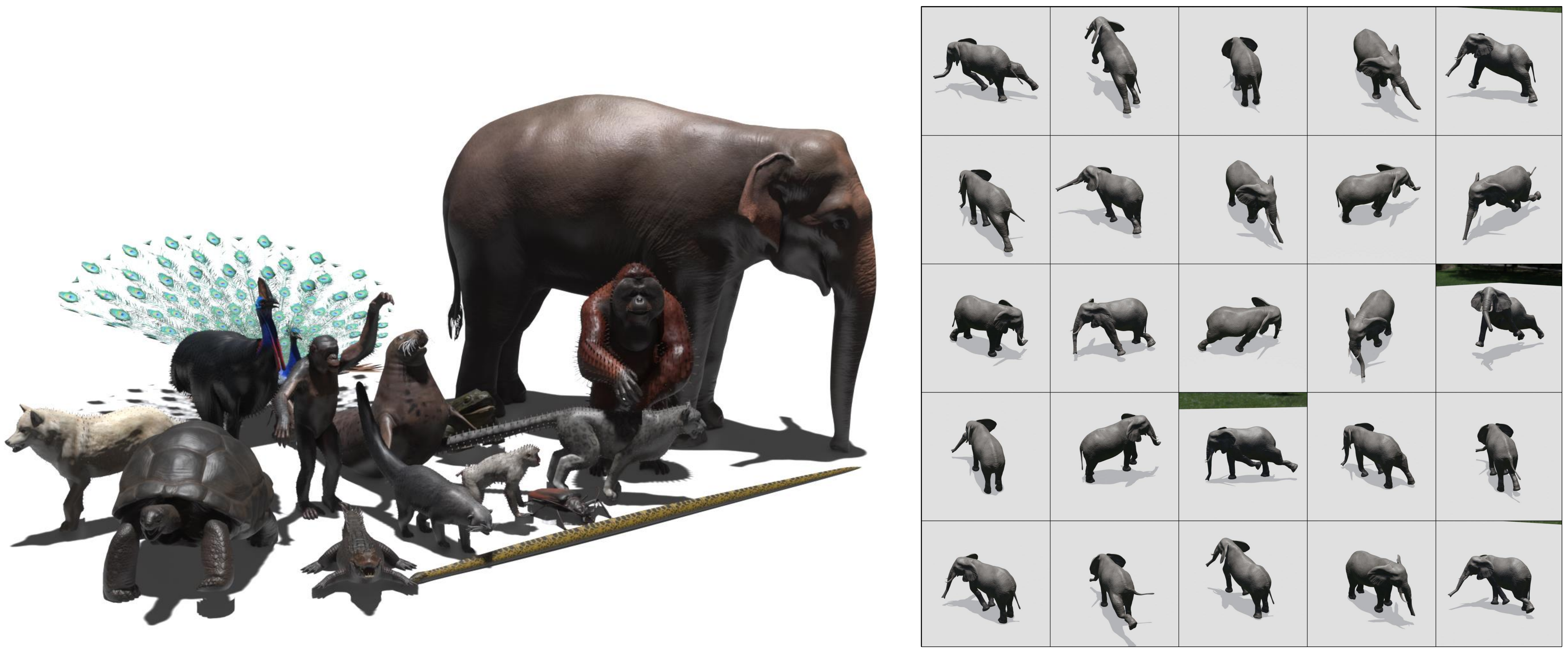}
\caption{\textbf{Overview of \dataset{}}. We collect a novel dataset \dataset{} containing high-fidelity and animatable 3D animal models. On the right we show renderings for the elephant with different poses.}
\label{fig:dataset}
\end{figure}

We provide the names of all categories in PlanetZoo in Tab.~\ref{tab:all_names}. We also show an overview figure of PlanetZoo in Fig.~\ref{fig:dataset}.

\clearpage
\begin{longtable}[h]{ccc}
\caption{Names of all categories in PlanetZoo.} 
\label{tab:all_names}
\\ \toprule
  aardvark female &   aardvark juvenile &   aardvark male \\
  african buffalo female &   african buffalo juvenile &   african buffalo male \\
  african elephant female &   african elephant juvenile &   african elephant male \\
  african wild dog female &   african wild dog juvenile &   african wild dog male \\
  aldabra giant tortoise female &   aldabra giant tortoise juvenile &   aldabra giant tortoise male \\
  amazonian giant centipede &   american bison female &   american bison juvenile \\
  american bison male &   arctic wolf female &   arctic wolf juvenile \\
  arctic wolf male &   babirusa female &   babirusa juvenile \\
  babirusa male &   bactrian camel female &   bactrian camel juvenile \\
  bactrian camel male &   bairds tapir female &   bairds tapir juvenile \\
  bairds tapir male &   bengal tiger female &   bengal tiger juvenile \\
  bengal tiger male &   binturong female &   binturong juvenile \\
  binturong male &   black wildebeest female &   black wildebeest juvenile \\
  black wildebeest male &   boa constrictor &   bongo female \\
  bongo male &   bonobo female &   bonobo juvenile \\
  bonobo male &   bornean orangutan female &   bornean orangutan juvenile \\
  bornean orangutan male &   brazilian salmon pink tarantula &   brazilian wandering spider \\
  capuchin monkey female &   capuchin monkey juvenile &   capuchin monkey male \\
  cassowary female &   cassowary juvenile &   cassowary male \\
  cheetah female &   cheetah juvenile &   cheetah male \\
  chinese pangolin female &   chinese pangolin juvenile &   chinese pangolin male \\
  clouded leopard female &   clouded leopard juvenile &   clouded leopard male \\
  common ostrich female &   common ostrich juvenile &   common ostrich male \\
  common warthog female &   common warthog juvenile &   common warthog male \\
  cuviers dwarf caiman female &   cuviers dwarf caiman juvenile &   cuviers dwarf caiman male \\
  dall sheep female &   dall sheep juvenile &   dall sheep male \\
  death adder &   dhole female &   dhole juvenile \\
  dhole male &   diamondback terrapin &   dingo female \\
  dingo juvenile &   dingo male &   eastern blue tongued lizard \\
  eastern brown snake &   formosan black bear female &   formosan black bear juvenile \\
  formosan black bear male &   galapagos giant tortoise female &   galapagos giant tortoise juvenile \\
  galapagos giant tortoise male &   gemsbok female &   gemsbok juvenile \\
  gemsbok male &   gharial female &   gharial juvenile \\
  gharial male &   giant anteater female &   giant anteater juvenile \\
  giant anteater male &   giant burrowing cockroach &   giant desert hairy scorpion \\
  giant forest scorpion &   giant leaf insect &   giant otter female \\
  giant otter juvenile &   giant otter male &   giant panda female \\
  giant panda juvenile &   giant panda male &   gila monster \\
  golden poison frog &   goliath beetle &   goliath birdeater \\
  goliath frog &   gray wolf female &   gray wolf juvenile \\
  gray wolf male &   greater flamingo female &   greater flamingo juvenile \\
  greater flamingo male &   green iguana &   grey seal female \\
  grey seal juvenile &   grey seal male &   grizzly bear female \\
  grizzly bear juvenile &   grizzly bear male &   himalayan brown bear female \\
  himalayan brown bear juvenile &   himalayan brown bear male &   hippopotamus female \\
  hippopotamus juvenile &   hippopotamus male &   indian elephant female \\
  indian elephant juvenile &   indian elephant male &   indian peafowl juvenile \\
  indian peafowl male &   indian rhinoceros female &   indian rhinoceros juvenile \\
  indian rhinoceros male &   jaguar female &   jaguar juvenile \\
  jaguar male &   japanese macaque female &   japanese macaque juvenile \\
  japanese macaque male &   king penguin female &   king penguin juvenile \\
  king penguin male &   koala female &   koala juvenile \\
  koala male &   komodo dragon female &   komodo dragon juvenile \\
  komodo dragon male &   lehmanns poison frog &   lesser antillean iguana \\
  llama female &   llama juvenile &   llama male \\
  malayan tapir female &   malayan tapir juvenile &   malayan tapir male \\
  mandrill female &   mandrill juvenile &   mandrill male \\
  mexican redknee tarantula &   nile monitor juvenile &   nile monitor male \\
  nyala female &   nyala juvenile &   nyala male \\
  okapi female &   okapi juvenile &   okapi male \\
  plains zebra female &   plains zebra juvenile &   plains zebra male \\
  polar bear female &   polar bear juvenile &   polar bear male \\
  proboscis monkey female &   proboscis monkey juvenile &   proboscis monkey male \\
  pronghorn antelope female &   pronghorn antelope juvenile &   pronghorn antelope male \\
  puff adder &   pygmy hippo female &   pygmy hippo juvenile \\
  pygmy hippo male &   red eyed tree frog &   red kangaroo female \\
  red kangaroo juvenile &   red kangaroo male &   red panda juvenile \\
  red panda male &   red ruffed lemur female &   red ruffed lemur juvenile \\
  red ruffed lemur male &   reindeer female &   reindeer juvenile \\
  reindeer male &   reticulated giraffe female &   reticulated giraffe juvenile \\
  reticulated giraffe male &   ring tailed lemur juvenile &   ring tailed lemur male \\
  sable antelope female &   sable antelope juvenile &   sable antelope male \\
  saltwater crocodile female &   saltwater crocodile juvenile &   saltwater crocodile male \\
  siberian tiger female &   siberian tiger juvenile &   siberian tiger male \\
  snow leopard female &   snow leopard juvenile &   snow leopard male \\
  spotted hyena female &   spotted hyena juvenile &   spotted hyena male \\
  springbok female &   springbok juvenile &   springbok male \\
  sun bear female &   sun bear juvenile &   sun bear male \\
  thomsons gazelle female &   thomsons gazelle juvenile &   thomsons gazelle male \\
  titan beetle &   west african lion female &   west african lion juvenile \\
  west african lion male &   western chimpanzee female &   western chimpanzee juvenile \\
  western chimpanzee male &   western diamondback rattlesnake &   western lowland gorilla female \\
  western lowland gorilla juvenile &   western lowland gorilla male &   yellow anaconda \\
\bottomrule
\end{longtable}

\begin{table*}[h]
\small
\centering
\caption{Full retrieval results.}
\label{tab:all_retrieval}
\begin{tabular}{cc}
\toprule
Input animal & Retrieval animal \\
\cmidrule(l){1-2}
aardvark female  &  koala female \\
aardvark juvenile  &  arctic wolf juvenile \\
aardvark male  &  babirusa juvenile \\
african elephant female  &  indian elephant male \\
african elephant male  &  indian elephant juvenile \\
african elephant juvenile  &  indian elephant juvenile \\
binturong female  &  giant otter male \\
binturong juvenile  &  cuviers dwarf caiman female \\
binturong male  &  cuviers dwarf caiman female \\
grey seal female  &  babirusa juvenile \\
grey seal juvenile  &  koala juvenile \\
grey seal male  &  giant otter female \\
bonobo juvenile  &  bornean orangutan juvenile \\
bonobo male  &  western chimpanzee juvenile \\
bonobo female  &  western chimpanzee juvenile \\
polar bear female  &  formosan black bear male \\
polar bear juvenile  &  spotted hyena juvenile \\
polar bear male  &  grizzly bear male \\
gray wolf female  &  spotted hyena female \\
gray wolf juvenile  &  arctic wolf juvenile \\
gray wolf male  &  spotted hyena male \\
common ostrich female  &  american bison juvenile \\
common ostrich juvenile  &  babirusa juvenile \\
common ostrich male  &  greater flamingo juvenile \\
\bottomrule
\end{tabular}
\end{table*}

\end{document}